\documentclass{article}
\usepackage{colortbl}
\usepackage{microtype}
\usepackage{graphicx}
\usepackage{subcaption}
\usepackage{booktabs} 
\usepackage{array}
\usepackage{multirow}
\usepackage{makecell}
\usepackage{xcolor}
\usepackage{pifont} 
\definecolor{special_yellow}{rgb}{1, 0.7412, 0.0353}
\definecolor{special_green}{rgb}{0.4314, 0.6392, 0.596}
\newcommand{\tick}{\textcolor{green}{\ding{51}}}  
\newcommand{\cross}{\textcolor{red}{\ding{55}}} 
\newcommand{\hollowstar}{\text{\ding{73}}}
\newcommand{\solidstar}{\text{\ding{72}}}
\usepackage{titletoc}
\usepackage{listings}
\usepackage[many,listingsutf8]{tcolorbox}
\usepackage[frozencache,cachedir=minted-cache]{minted} 

\lstdefinelanguage{json}{
    basicstyle=\ttfamily\footnotesize,
    numbers=left,
    numberstyle=\tiny\color{gray},
    stepnumber=1,
    numbersep=5pt,
    showstringspaces=false,
    breaklines=true,
    frame=single,
    backgroundcolor=\color{gray!10},
    literate=
     *{0}{{{\color{blue}0}}}{1}
      {1}{{{\color{blue}1}}}{1}
      {2}{{{\color{blue}2}}}{1}
      {3}{{{\color{blue}3}}}{1}
      {4}{{{\color{blue}4}}}{1}
      {5}{{{\color{blue}5}}}{1}
      {6}{{{\color{blue}6}}}{1}
      {7}{{{\color{blue}7}}}{1}
      {8}{{{\color{blue}8}}}{1}
      {9}{{{\color{blue}9}}}{1},
}

\usepackage[colorlinks=true, linkcolor=red, citecolor=blue, urlcolor=blue]{hyperref}

\usepackage[accepted]{icml2025}

\usepackage{amsmath}
\usepackage{amssymb}
\usepackage{mathtools}
\usepackage{amsthm}
\usepackage{algorithm}
\usepackage{algorithmic}

\usepackage{tocloft}
\usepackage{titlesec}

\renewcommand{\cftsecfont}{\bfseries}
\renewcommand{\cftsubsecfont}{\normalfont}

\renewcommand{\cftsecpagefont}{\normalfont}

\usepackage[capitalize,noabbrev]{cleveref}

\theoremstyle{plain}

\theoremstyle{definition}

\theoremstyle{remark}

\usepackage[textsize=tiny]{todonotes}
\usepackage{marvosym}

\icmltitlerunning{A Scalable Multi-Dimensional Benchmark for Essential Virtual Agent Capabilities}

\begin{document}
\addtocontents{toc}{\protect\setcounter{tocdepth}{-1}}

\twocolumn[
\icmltitle{What Limits Virtual Agent Application? \\ OmniBench: A Scalable Multi-Dimensional Benchmark for Essential Virtual Agent Capabilities}



\icmlsetsymbol{equal}{*}
\icmlsetsymbol{corr1}{$\dagger$}
\icmlsetsymbol{corr2}{\Letter}

\begin{icmlauthorlist}
\icmlauthor{Wendong Bu}{zju,ant,equal}
\icmlauthor{Yang Wu}{ant,equal}
\icmlauthor{Qifan Yu}{zju,equal}
\icmlauthor{Minghe Gao}{zju}
\icmlauthor{Bingchen Miao}{zju}
\icmlauthor{Zhenkui Zhang}{zju}
\icmlauthor{Kaihang Pan}{zju}
\icmlauthor{Yunfei Li}{ant}
\icmlauthor{Mengze Li}{hkust}
\icmlauthor{Wei Ji}{nju}
\icmlauthor{Juncheng Li}{zju,corr2}
\icmlauthor{Siliang Tang}{zju}
\icmlauthor{Yueting Zhuang}{zju}
\end{icmlauthorlist}

\icmlaffiliation{zju}{Zhejiang University, Hangzhou, China}
\icmlaffiliation{ant}{Ant Group, Hangzhou, China}
\icmlaffiliation{hkust}{The Hong Kong University of Science and Technology, Hong Kong SAR, China}
\icmlaffiliation{nju}{Nanjing University, Nanjing, China}

\icmlcorrespondingauthor{Juncheng Li}{junchengli@zju.edu.cn}

\icmlkeywords{Machine Learning, ICML}

\vskip 0.3in
]



\printAffiliationsAndNotice{\icmlEqualContribution} 

\begin{abstract}

As multimodal large language models (MLLMs) advance, MLLM-based virtual agents have demonstrated remarkable performance. However, existing benchmarks face significant limitations, including uncontrollable task complexity, extensive manual annotation with limited scenarios, and a lack of multidimensional evaluation. In response to these challenges, we introduce \textbf{\mbox{OmniBench}}, a self-generating, cross-platform, graph-based benchmark with an automated pipeline for synthesizing tasks of controllable complexity through subtask composition. To evaluate the diverse capabilities of virtual agents on the graph, we further present \textbf{OmniEval}, a multidimensional evaluation framework that includes subtask-level evaluation, graph-based metrics, and comprehensive tests across 10 capabilities. Our synthesized dataset contains 36k graph-structured tasks across 20 scenarios, achieving a 91\% human acceptance rate. Training on our graph-structured data shows that it can more efficiently guide agents compared to manually annotated data. We conduct multidimensional evaluations for various open-source and closed-source models, revealing their performance across various capabilities and paving the way for future advancements. Our project is available at \mbox{\url{https://omni-bench.github.io/}}.

\end{abstract}

\vspace{-10mm}
\section{Introduction}
\vspace{-2mm}

\begin{figure*}[h!]
\vspace{-3mm}
\includegraphics[width=\linewidth]{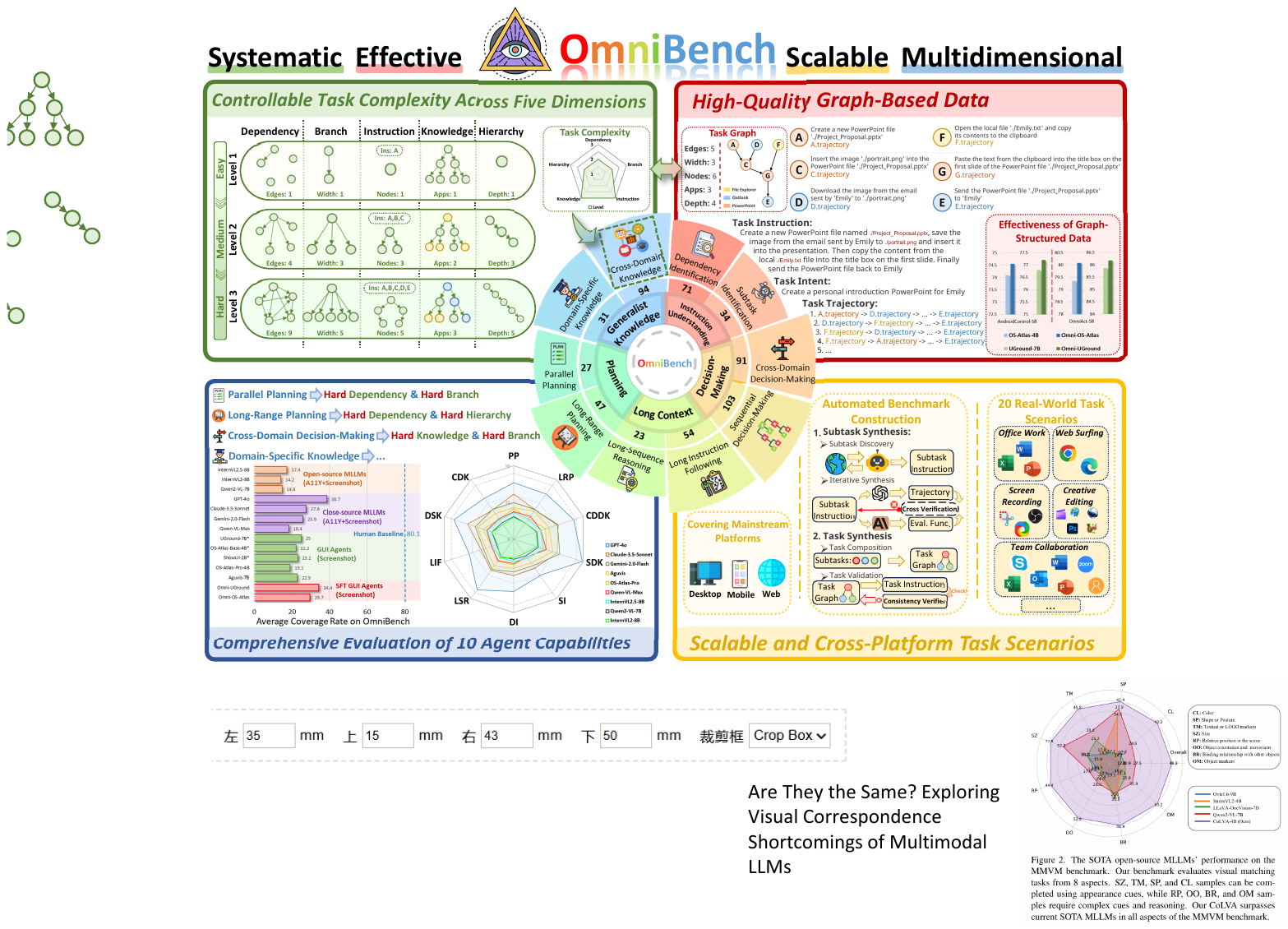}
\vspace{-10mm}
\centering\caption{Overview of OmniBench, a systematic benchmark with five-dimensional task complexity and bottom-up automatic task synthesis for generating structured task graphs. It evaluates ten virtual agent capabilities using high-quality graph-based data, ensuring scalable and realistic task evaluation.}
\label{mainfigure}
\vspace{-6mm}
\end{figure*}
With the development of MLLMs~\cite{fei2024vitron, wu24next}, recent MLLM-based virtual agents~\cite{aguvis, gva, miao2025boostingvirtualagentlearning} have demonstrated promising performance in web navigation~\cite{scribeagent}, mobile device control~\cite{iris}, and computer interaction~\cite{dawn}. 
To explore real-world values of visual agents, current research mainly evaluates their performance based on offline trajectory similarity with human demonstrations~\cite{aitw, gui-odyssey, mind2web} or by using expert-crafted functions in interactive online environments~\cite{osworld, webarena, webshop}.

However, these two types of benchmarks mentioned above still have notable limitations: 
\textbf{1)~Uncontrollable and fixed task complexity.} 
Existing benchmarks typically propose tasks entirely rather than progressively with fine-grained guidance,  which results in uncontrollable and fixed task complexity. Uncontrollable task complexity makes it hard to design fine-grained test data for various capabilities, while fixed complexity makes it challenging for benchmarks to keep up with agents' growing capabilities. \textbf{2)~Extensive manual labor and limited task scenarios.} The existing benchmarks rely on manual annotations to synthesize demonstration trajectories or evaluation functions, making the cost of designing benchmarks unaffordable and hindering the expansion of scale. Moreover, the annotated data with a limited amount is influenced by human prior experience, making it difficult to cover comprehensive scenarios.
\textbf{3)~Absence of multidimensional evaluation}: Existing benchmarks commonly evaluate agents based on the final state of tasks, lacking an evaluation of the intermediate steps in task execution. Additionally, the various capabilities required by virtual agents to complete tasks (\textit{e.g.}, planning, instruction understanding, etc.) cannot be quantified by coarse task success rates, failing to provide sufficient feedback for potential future improvements. \textbf{In summary}, an ideal benchmark should include not only diverse task scenarios with controllable complexity, but also a comprehensive evaluation across multiple dimensions.

To cost-effectively construct diverse task scenarios with complexity at multiple granularities for comprehensive agent evaluation, we propose a novel self-generating, graph-based benchmark, \textbf{\mbox{OmniBench}}. It dynamically synthesizes tasks with controllable complexity based on a bottom-up pipeline.
OmniBench spans five fundamental types of task complexity to construct 10 evaluation dimensions (see \mbox{Figure~\ref{mainfigure}}). Test tasks across these dimensions are categorized based on combinations of complexity types. For example, a long-range planning test task typically exhibits higher dependency complexity and hierarchy complexity.
OmniBench consists of 36k high-quality graph-structured tasks across 20 distinct scenarios~(\textit{e.g.} image editing, video editing) derived from its self-generating framework, with the task scale being 40x larger than most environment-based benchmarks, as shown in Table~\ref{tab1}.
This automated process opens up the possibility of scaling up virtual agent evaluation in a low-resource manner. Therefore, OmniBench facilitates the easy construction of agent benchmarks on desktop, mobile, and web platforms, as shown in Table~\ref{tab1}. For multidimensional task synthesis, we take motivations from the DAG topology~\cite{crab} to design a bottom-up pipeline. Specifically, we define five fundamental task complexities and \textbf{synthesize tasks with controllable complexity by constraining the DAG composition process}. Notably, we extracted task intents to guide this process to avoid
composing meaningless tasks (\textit{e.g.}, opening a food delivery app and immediately closing it). We further incorporate quality control modules to optimize subtasks and ensure semantic alignment. In this way, we cost-effectively synthesize high-quality graph-structured tasks, significantly \textbf{broadening task scenarios without requiring human annotations}.

\begin{table*}[h!]
\vspace{-6mm}
\caption{Comparison of virtual agent benchmarks across environment, task, and evaluation dimensions. Unlike previous benchmarks, OmniBench features automatic task composition, five-dimensional task complexity, and a 10-capability evaluation framework.}
\centering
\resizebox{\textwidth}{!}{%
\begin{tabular}{l|ccc|cccccccc|cccc}
\toprule
\multirow{2}{*}{} & \multicolumn{3}{c|}{\textbf{Environment}} & \multicolumn{8}{c|}{\textbf{Task}} & \multicolumn{4}{c}{\textbf{Evaluation}} \\ \cline{2-16}
& \textbf{\makecell{Interactive}} & \textbf{\makecell{Real-World}} & \textbf{\makecell{Platform}} & \textbf{\makecell{\# Instance}} & \textbf{\makecell{\# Compl.\\Dimen.}} & \textbf{\makecell{Dyna.\\Scale}} & \textbf{Intent} & \textbf{\# Scenario} & \textbf{\makecell{Demo.\\Traj.}} & \textbf{Construction} & \textbf{\makecell{Instruction\\Level}} & \textbf{\makecell{\# Cap.\\Dimen.}} & \textbf{\makecell{Eval.\\Level}} & \textbf{\makecell{\# Eval.\\Func.}} & \textbf{\makecell{Evaluation\\Strategy}} \\ \hline
 AitW~\cite{aitw} & \cross & \cross & \includegraphics[width=0.4cm]{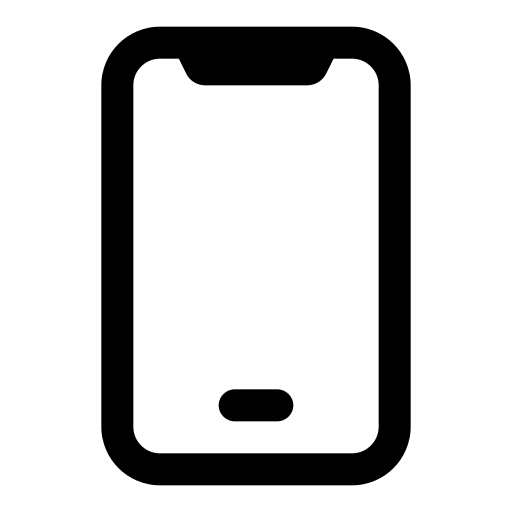} & 30378 & - & \cross & \cross & 5 & \tick & Manual Annotation & High \& Low & 1 & Task & - & Trajectory-based \\
 Mind2Web~\cite{mind2web} & \cross & \cross & \includegraphics[width=0.4cm]{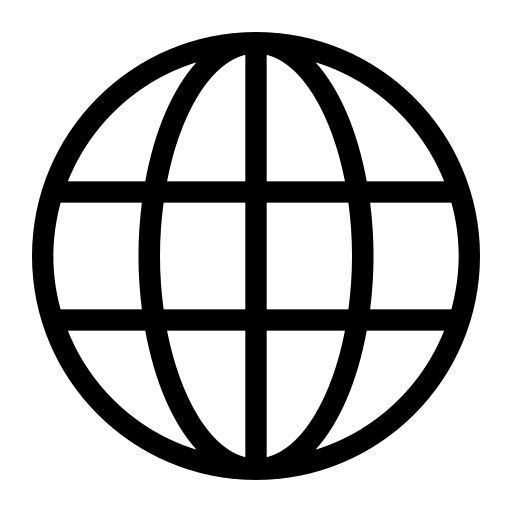} & 2350 & - & \cross & \cross & 5 & \tick & Manual Annotation & High & 1 & Task & - & Trajectory-based \\ 
 MoTIF~\cite{motif} & \cross & \cross & \includegraphics[width=0.4cm]{assets/mobile.png} & 756 & - & \cross & \cross & - & \tick & Manual Annotation & High \& Low & 1 & Task & - & Trajectory-based \\
 OmniACT~\cite{omniact} & \cross & \cross & \includegraphics[width=0.4cm]{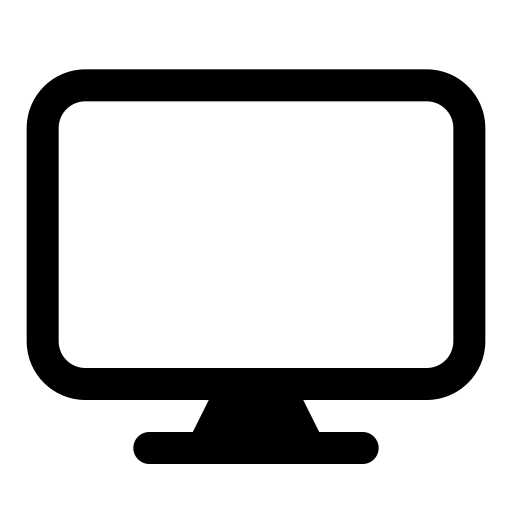} \includegraphics[width=0.4cm]{assets/web.png} & 9802 & - & \cross & \cross & 6 & \tick & Manual Annotation & Low & 1 & Task & - & Trajectory-based \\
 GUI Odyssey~\cite{gui-odyssey} & \cross & \cross & \includegraphics[width=0.4cm]{assets/mobile.png} \includegraphics[width=0.4cm]{assets/web.png} & 7735 & - & \cross & \cross & 6 & \tick & Manual Annotation & Low & 1 & Task & - & Trajectory-based \\ \hline
 WebArena~\cite{webarena} & \tick & \cross & \includegraphics[width=0.4cm]{assets/web.png} & 812 & - & \cross & \cross & 4 & \cross & Manual Annotation & Low & 1 & Task & 5 & Result-based \\
 VisualWebArena~\cite{visualwebarena} & \tick & \cross & \includegraphics[width=0.4cm]{assets/web.png} & 910 & 2 & \cross & \cross & 3 & \cross & Manual Annotation & Low & 1 & Task & 6 & Result-based \\ 
 OSWorld~\cite{osworld} & \tick & \tick & \includegraphics[width=0.4cm]{assets/desktop.png} & 369 & - & \cross & \cross & 5 & \tick & Manual Annotation & Low & 1 & Task & 134 & Result-based \\
 Spider2-V~\cite{spider2-v} & \tick & \tick & \includegraphics[width=0.4cm]{assets/desktop.png} \includegraphics[width=0.4cm]{assets/web.png} & 494 & 1 & \cross & \cross & 7 & \cross & Manual Annotation & High \& Low & 1 & Task & 151 & Result-based \\ 
 CRAB~\cite{crab} & \tick & \tick & \includegraphics[width=0.4cm]{assets/desktop.png} \includegraphics[width=0.4cm]{assets/mobile.png} & 100 & - & \tick & \cross & - & \cross & Manual Composition & Low & 1 & Subtask & 59 & Graph-based \\ \hline
 OmniBench~(Ours) & \tick & \tick & \includegraphics[width=0.4cm]{assets/desktop.png} \includegraphics[width=0.4cm]{assets/mobile.png} \includegraphics[width=0.4cm]{assets/web.png} & 36076 & 5 & \tick & \tick & 20 & \tick & \makecell{Automatic Composition \\ \& Human Verification} & High \& Low & 10 & Subtask & 255 & Graph-based

 \\ \bottomrule
\end{tabular}%
}
\label{tab1}
\vspace{-6mm}
\end{table*}

Moreover, due to the inherent complexity of conducting comprehensive and fine-grained evaluations of agents, we propose a graph-based multidimensional evaluation framework, \textbf{OmniEval}. In contrast to previous coarse-grained evaluation methods, we introduce a graph-based evaluator that leverages subtask-level evaluation functions synthesized in OmniBench. Specifically, we design two novel fine-grained metrics to evaluate agents' performance on graph-structured tasks and their alignment with human logic.
Based on \mbox{OmniBench}, we comprehensively evaluate 12 virtual agents, including both open-source and proprietary models, across all 10 capability dimensions as shown in Figure~\ref{mainfigure}, \textbf{fully revealing the capability boundaries and providing concrete directions for future improvement}.

The performance comparison between various models provides valuable findings for future virtual agent application. Specifically:
\textbf{1)~Existing agents struggle to handle graph-structured tasks}. 
Compared to tasks with linear structures, the agents fall significantly short when facing graph-structured tasks, with even GPT-4o achieving only 20.5\% performance, while humans can reach 80.1\%. 
\textbf{2)~Task intents are crucial for task planning}. 
Incorporating task intents into the prompt offers a plug-and-play improvement to planning performance, with an average increase from 23.4\% to 28.9\%. Similarly, using task intents in fine-tuning data improves planning performance from 30.5\% to 31.9\%.
\textbf{3)~Mainstream agents are sensitive to expression order in task instructions}. 
We observe significant performance fluctuations in existing agents when altering the expression order of task instructions. In contrast, agents fine-tuned with graph-structured trajectories exhibit more stable performance.

Furthermore, we fine-tune two open-source agents with distinct architectures on synthesized graph-structured task trajectories. As shown in Figure~\ref{mainfigure}, both agents exhibit performance improvements on AndroidControl and OmniAct. Compared to agents trained on manually annotated datasets, our agents achieve better performance across diverse benchmarks, benefiting from reasoning-rich trajectories that demonstrate their broad applicability and strong potential.

\vspace{-3mm}
\section{Related Work}
\vspace{-2mm}
\textbf{Virtual Digital Agents.}
With the development of MLLMs~\cite{pan2025generative, pan2024i3, li2023fine, fei2024dysen}, virtual agents have greatly improved task automation across platforms. CogAgent~\cite{cogagent} introduced an 18B visual language model for GUI understanding, achieving state-of-the-art performance. SeeClick~\cite{seeclick} developed a vision-only model that interacts with GUIs via screenshots, eliminating the need for structured data. UGround~\cite{uground} proposed a universal grounding model, accurately mapping GUI elements across platforms. Iris~\cite{iris} enhances GUI automation by tackling challenges in complex digital environments. Evaluating these visual agents is crucial for real-world applications.

\textbf{Benchmarks for Virtual Agents.} 
Mainstream benchmarks for virtual agents are generally categorized into two types: trajectory-based and result-based. Trajectory-based benchmarks \cite{aitw,seeclick,mind2web} compare agent trajectories to human demonstrations but can be inaccurate due to the existence of multiple valid trajectories. Result-based benchmarks \cite{osworld,webarena,spider2-v} focus on the final state of the environment, overlooking the fine-grained evaluation of intermediate processes.
More recently, some studies \cite{taskbench,crab} have introduced graph-based evaluations, which support both multiple feasible trajectories and the evaluation of intermediate processes. TASKBENCH~\cite{taskbench} evaluates agents for task automation, but its simplistic metrics fail to fully utilize the potential of graph structures. CRAB~\cite{crab} evaluates agents using handcrafted graphs, but it lacks a systematic task analysis, limiting fine-grained capability assessment. Notably, to the best of our knowledge, \textbf{\mbox{OmniBench}} is the only scalable benchmark for virtual agents that defines composable task complexity using graphs to evaluate multiple essential capabilities.

\vspace{-3mm}
\section{OmniBench}
\vspace{-2mm}

\begin{figure*}[h!]
\vspace{-3mm}
\includegraphics[width=\linewidth]{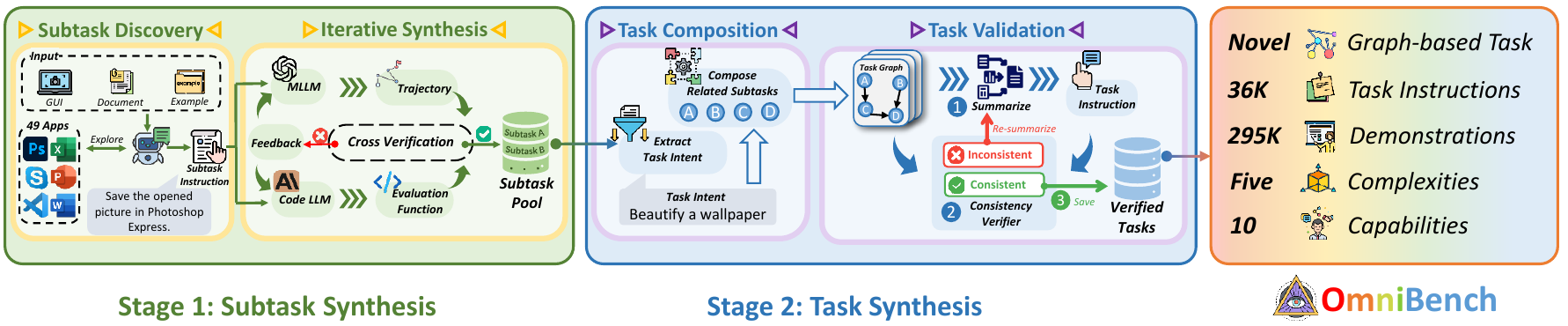}
\vspace{-10mm}
\centering\caption{Bottom-up task synthesis pipeline}
\label{pipeline}
\vspace{-6mm}
\end{figure*}

\textls[-30]{OmniBench consists of 36k high-quality graph-structured tasks across 10 evaluation dimensions to simulate the way humans perceive the digital world, including planning, decision-making, etc. In this section, we first introduce task graphs of OmniBench and systematically define corresponding task complexities~\mbox{(Section \ref{sec:3.1})}. Then, we present the bottom-up data collection pipeline for controllably synthesizing tasks~\mbox{(Section \ref{sec:3.2})}. Additionally, we explain how to control the quality of the synthesized data~\mbox{(Section \ref{sec:3.3})}. Finally, we showcase the statistics of OmniBench~\mbox{(Section \ref{sec:3.4})}.}

\vspace{-2mm}
\subsection{Task Complexity on Task Graph}
\vspace{-2mm}
\label{sec:3.1}

\textls[-30]{In this section, we define task complexity on graphs, which is later constrained in Section~\ref{sec:4.2} to construct test tasks for multidimensional capability. We propose a new complexity definition because existing benchmarks~\cite{mobileagentbench,spider2-v,visualwebarena} typically define task complexity based on the number of steps in human demonstrations. However, this approach has two limitations: 1)~The inherent subjectivity of human demonstrations makes this definition unreliable; 2)~It is one-dimensional and fails to capture the multifaceted complexity of real-world tasks. Inspired by previous works~\cite{crab,taskbench} that represent tasks as graph structures, we introduced the concept of the task graph and systematically defined five fundamental task complexities on the task graph.}

\begin{table}[t!]
\centering
\vspace{-3mm}
\caption{Complexity Dimensions and Their Corresponding Levels}
\resizebox{\linewidth}{!}{
\begin{tabular}{ccccc}
\toprule
\textbf{Complexity Dimension} & \textbf{Calculation} & \textbf{Easy} & \textbf{Medium} & \textbf{Hard} \\ \midrule
Dependency Complexity & Number of Edges & $\leq 1$ & $2 \sim 3$ & $\geq 4$ \\ 
Instruction Complexity & Number of Nodes & $\leq 2$ & $3 \sim 4$ & $\geq 5$ \\ 
Knowledge Complexity & Number of Application Categories & $\leq 1$ & $2 \sim 3$ & $\geq 4$ \\ 
Hierarchy Complexity & Depth & $\leq 2$ & $3 \sim 4$ & $\geq 5$ \\ 
Branch Complexity & Width & $\leq 2$ & $3 \sim 4$ & $\geq 5$ \\ \bottomrule
\end{tabular}
}
\vspace{-9mm}
\label{tab:complexity}
\end{table}

\textls[-30]{Specifically, we define a subtask as a smaller, independent task that contributes to completing a more complex task. Each subtask has input and output resources to constrain the dependencies between them. To formalize this, we assume each subtask as \( s \) and define a task graph as \( \mathcal{G} = \{S, R\} \), where \( S = \{s_1, s_2, \dots, s_n\} \) is the collection of subtasks, and \( R \) is a set of relations \( \{(s_a, s_b)\} \) indicating that subtask \( s_b \) depends on subtask \( s_a \) when the output resources of \( s_a \) match the input resources of \( s_b \).}

\textls[-30]{After representing the task as a graph, a natural idea is to define task complexity based on the topology of the task graph. We systematically analyzed five characteristics of the graph and designed a corresponding five-dimensional task complexity. Specifically: 
\textbf{1)~Dependency Complexity.} Since each edge in the task graph represents a dependency between subtasks, we define dependency complexity based on the number of edges. 
\textbf{2)~Instruction Complexity.} The semantics of a task instruction are composed of all subtasks, with more subtasks leading to more complex instruction semantics. Therefore, we define instruction complexity based on the number of nodes. 
\textbf{3) Knowledge Complexity.} We categorize all 49 applications and define knowledge complexity based on the number of applications from different categories (\textit{e.g.}, multimedia playback, productivity) in the task graph. The detailed categorization is provided in Appendix\textcolor{red}{\hypersetup{linkcolor=red}\hyperref[app_cate]{~\ref{app_cate}}}.
\textbf{4)~Hierarchy Complexity.} The depth of the task graph represents the number of hierarchical levels in the task structure. Thus, we define hierarchy complexity based on the depth. 
\textbf{5)~Branch Complexity.} A wider task graph indicates more branches that can be executed concurrently. Therefore, we define branch complexity based on the width. 
The classification criteria for three complexity levels are shown in Table~\ref{tab:complexity}.}

\vspace{-2mm}
\subsection{Controllable Task Synthesis}
\vspace{-2mm}
\label{sec:3.2}
\textls[-30]{Although we define five fundamental task complexities on the task graph, converting task instructions into a graph remains challenging. A straightforward idea is to directly top-down decompose tasks into task graphs. However, this process is typically based on uncontrollable MLLMs or expensive manual efforts. Therefore, to effectively synthesize tasks with controllable complexity, we designed a bottom-up automated task synthesis pipeline, as shown in Figure~\ref{pipeline}.}

\textls[-30]{\noindent \textbf{Overview.} The task synthesis pipeline we propose consists of four processes. First, we synthesize a series of simple subtask instructions from the explorable environment. Then, we iteratively synthesize subtask trajectories and evaluation functions. Next, the subtasks are combined into a task bottom-up. Finally, we validate the semantics of the tasks.}

\begin{table}[t!]
\vspace{-3.5mm}
\centering
\caption{Ablation study evaluating the impact of each quality control module on acceptability of the synthesized tasks.}
\resizebox{\linewidth}{!}{
\begin{tabular}{ccc|c}
\toprule
\textbf{Cross-Verification} & \textbf{Intent Extraction} & \textbf{Consistency Validator} & \multicolumn{1}{c}{\mbox{\textbf{Human Acceptance}}} \\ 
\midrule
 \cross & \cross & \cross & 41.2\% \\ 
 \cross & \tick & \tick & 61.2\% \\ 
 \tick & \cross & \tick & 82.7\% \\ 
 \tick & \tick & \cross & 86.5\% \\ 
 \tick & \tick & \tick & 90.7\% \\ 
 \bottomrule
\end{tabular}
}
\label{ablation}
\vspace{-9mm}
\end{table}

\textls[-30]{\noindent \textbf{Subtask Exploration.} 
We designed an environment containing 49 diverse applications, inspired by OSWorld~\cite{osworld}, allowing advanced MLLMs to thoroughly explore each application to propose diverse and achievable subtasks. During exploration, documentation and example subtasks for each application are provided to help synthesis. To accurately synthesize the dependencies between subtasks, we provide a predefined resource list, which MLLMs use to determine the input and output resources for each subtask. The implementation details can be found in Appendix\textcolor{red}{\hypersetup{linkcolor=red}\hyperref[A3.2.1]{~\ref{A3.2.1}}}.}

\textls[-30]{\noindent \textbf{Iterative Synthesis.} 
We leverage advanced MLLMs to synthesize trajectories and evaluation functions of subtasks. For trajectories, our crafted prompts guide the MLLM to describe screenshots and output thoughts, improving inference on trajectories. For evaluation functions, we predefine 11 basic APIs to retrieve information such as clicked text, keyboard inputs, and file directory existence. Subsequently, we use Claude-3.5-Sonnet, which excels in the code domain, to compose these basic APIs into evaluation functions for subtasks. To improve the quality of synthetic data, we propose a novel cross-verification algorithm that iteratively refines the synthesis process. The implementation details can be found in Appendix\textcolor{red}{\hypersetup{linkcolor=red}\hyperref[cross]{~\ref{cross}}}.}

\textls[-30]{\noindent \textbf{Task Composition.} For high-quality subtask samples that pass cross-verification, we add them to the subtask pool. Directly bottom-up composing these subtasks into a task graph using input and output resources may result in tasks that lack a coherent core goal, such as ``opening a food delivery app and immediately closing it". To avoid synthesizing such low-quality tasks, we extract task intents for the composition scenarios from the subtask pool, such as ``create a personal introduction PowerPoint for Emily," as shown in Figure~\ref{mainfigure}. Each task intent involves a group of subtasks, which are then combined into a task graph using input and output resources. Since the bottom-up composition process is rule-based, The synthesis of the task graph is controllable, ensuring that tasks with controllable complexity can be synthesized. Implementation details are in Appendix\textcolor{red}{\hypersetup{linkcolor=red}\hyperref[A3.2.3]{~\ref{A3.2.3}}}.}

\textls[-30]{\noindent \textbf{Task Validation.} For the task graphs constructed through composition, we employ GPT-4o to summarize the task instruction based on the subtask instructions and the graph structure. However, such synthesized instructions may deviate from the original semantics of the task graph, such as losing the graph's nonlinear semantics and degenerating into a simple sequential task. To ensure the synthesis of high-quality graph-structured tasks, we designed a consistency validator to verify the semantic alignment between the task graph and the summarized task instruction. Specifically, GPT-4o determine the dependency for subtasks solely based on the task instruction. If the inferred dependencies match those in the task graph, the instruction passes validation; otherwise, the instruction needs to be re-summarized. Implementation details are in Appendix\textcolor{red}{\hypersetup{linkcolor=red}\hyperref[A3.2.4]{~\ref{A3.2.4}}}.}

\vspace{-2mm}
\subsection{Quality Control}
\vspace{-2mm}
\label{sec:3.3}

\textls[-30]{Since the quality of graph-structured tasks is critical to the accurate evaluation of the virtual agents, we further introduce three designs to enhance the quality of synthesized data: a cross-verification mechanism, an intent extraction module, and a consistency validator. The cross-verification mechanism iteratively optimizes the demonstration trajectories and evaluation functions of subtasks, the intent extraction module ensures that the tasks have coherent goals, and the consistency validator aligns the semantics of the task graph and task instructions. We perform an ablation analysis of these three quality control methods, validating their effectiveness, as shown in Table~\ref{ablation}. 
For each ablation shown in the table, we sampled 400 task graphs and calculated the average acceptance by three specially trained annotators. The experiment shows that removing any quality control module decreases human acceptance, with the removal of the cross-verification algorithm resulting in the largest drop to 61.2\%.}

\vspace{-2mm}
\subsection{OmniBench Statistics}
\vspace{-2mm}
\label{sec:3.4}

\begin{figure}[htbp]
\vspace{-1.5mm}
    \centering
    \begin{minipage}[b]{0.4\textwidth}
        \centering
        \resizebox{\linewidth}{!}{%
        \begin{tabular}{lc}
        \toprule
        \textbf{Statistics} & \textbf{Value} \\ \midrule
        \textbf{Total Tasks} & 36076~(100\%) \\ 
        - Network-dependent Real-world Tasks & 16614~(46.05\%) \\
        - Network-independent Local Tasks & 19462~(53.95\%) \\ 
        - Avg. Number of Used App Per Task & 2.21 \\
        \midrule
        \textbf{Task Instruction} &  \\ 
        - Avg. Words of High-level Instruction & 51.7 \\
        - Avg. Words of Low-level Instruction & 237.9 \\ \midrule
        Total Task Scenarios & 20 \\ \midrule
        Total Subtasks & 255 \\ 
        \bottomrule
        \end{tabular}
        }
        \label{fig:statistics}
    \end{minipage}
    \hspace{0.05\textwidth}
    \begin{minipage}[b]{0.45\textwidth}
        \centering
        \includegraphics[width=\linewidth]{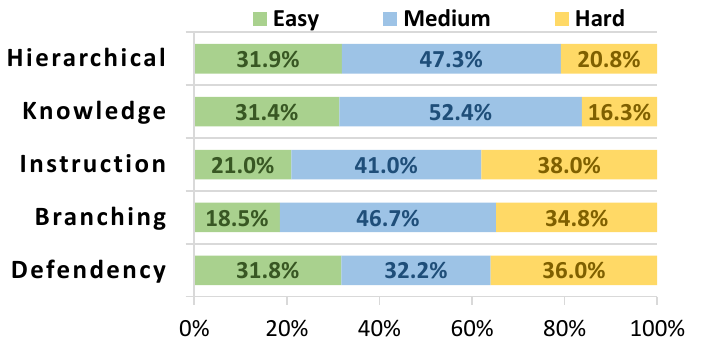}
        \vspace{-9mm}
        \caption{Statistics of OmniBench}
        \label{fig:complexity_distribution}
    \end{minipage}
    
    \vspace{-9mm}
    \label{fig:omni_overview}
\end{figure}

\textls[-30]{OmniBench comprises a total of 36,076 task instances, spanning across 20 common interactive scenarios and involving 49 diverse applications, as illustrated in Appendix\textcolor{red}{\hypersetup{linkcolor=red}\hyperref[app_cate]{~\ref{app_cate}}}. In Section~\ref{sec:3.1}, we introduced the five-dimensional complexity metrics for each task, with the distribution of different complexity levels for each dimension shown in Figure~\ref{fig:complexity_distribution}.}

\vspace{-3mm}
\section{OmniEval}
\vspace{-2mm}
\textls[-30]{To comprehensively analyze the limited capabilities of existing agents, we further propose OmniEval, a graph-based multidimensional evaluation framework. In this section, we first introduce a graph-based evaluator with two novel metrics for fine-grained and diverse evaluation~(Section~\ref{sec:4.1}). Then, we describe the construction of test tasks designed to evaluate 10 distinct capabilities by constraining task complexity~(Section~\ref{sec:4.2}).}

\vspace{-2mm}
\subsection{Graph-based Evaluator}
\vspace{-2mm}
\label{sec:4.1}

\begin{figure}[t]
\vspace{-2mm}
\includegraphics[width=\linewidth]{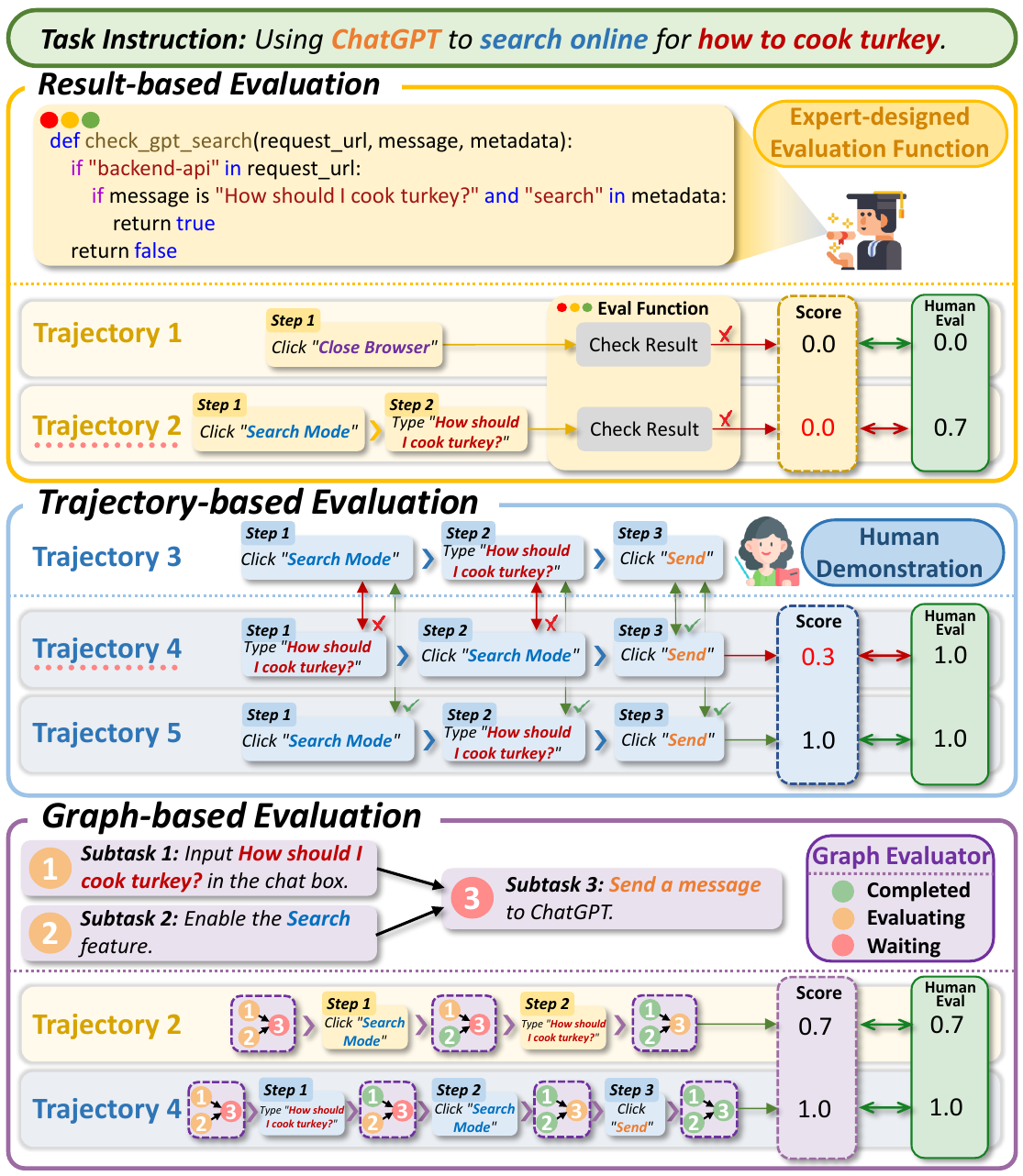}
\vspace{-8mm}
\centering\caption{Comparison of mainstream virtual agent evaluation strategies with the evaluation strategy we propose.}
\label{eval}
\vspace{-6mm}
\end{figure}

\textls[-30]{Currently, most benchmarks still evaluate agents in a coarse-grained and unreasonable paradigm. Specifically, \textbf{result-based} evaluations \cite{osworld,webarena} consider whether the final environment state aligns with expectations, lacking fine-grained intermediate evaluation. As shown in Figure~\ref{eval}, although both trajectories ultimately fail to complete the task, the progress of trajectory 2 is superior to trajectory 1, and they should not simply be categorized as failures. While \textbf{trajectory-based} evaluations \cite{aitw,gui-odyssey} compare the agents' predicted actions to human demonstrations at each step, they overlook multiple feasible trajectories. As shown in Figure~\ref{eval}, both trajectories can accomplish the task, but trajectory 4 is deemed a failure because it does not align with the demonstration trajectory, which is unreasonable.}

\textls[-30]{Considering the limitations of these two evaluation paradigms, we introduce a \textbf{graph-based} multi-metric evaluator inspired by previous research~\cite{crab}, as shown in Figure~\ref{eval}. Specifically, we define three evaluation states for each node on the task graph: Completed, Evaluating, and Waiting. Initially, when the evaluator is set up, nodes with an in-degree of 0 are marked as Evaluating, while the remaining nodes are marked as Waiting. After the agent executes each action, it checks whether all Evaluating nodes have been completed. Once a node is completed, it is marked as Completed, and new Evaluating nodes are added according to the topological order. We set a maximum number of steps $N$, and if the agent does not complete any subtasks within $N$ steps, the entire task is considered a failure.}

\textls[-30]{Additionally, to fully leverage the potential of the graph-based evaluator, we have designed two novel graph-based metrics. Traditional metrics fail to evaluate intermediate processes and alignment with human operational logic. For example, common metrics such as \textbf{Success Rate~(SR)}~\cite{osworld} focus on task outcomes rather than the process, while \textbf{Action Match Score (AMS)}~\cite{ams} treats action sequences as strings and compares the similarity between human demonstrations and agent predictions, rather than their logical similarity. To comprehensively quantify agent performance on the task graph, we propose two novel metrics inspired by its topology. The \textbf{Coverage Rate~(CR)} assesses agent progress on the task graph, while the \textbf{Logical Consistency~(LC)} reflects the similarity in operational logic between agents and humans.}

\textls[-30]{\textbf{Coverage Rate~(CR).} It evaluates an agent's progress on a task graph by weighting subtasks based on their depth, where deeper subtasks are assigned higher weights due to their increased number of prerequisite subtasks. Referring to the relevant definitions in Section~\ref{sec:3.1}, let \( d(s_i) \) denote the depth of subtask \( s_i \). The weight \( w(s_i) \) is:

\vspace{-7mm}

\[
w(s_i) = \frac{d(s_i)}{\sum_{j=1}^{n} d(s_j)}.
\]

\vspace{-4mm}

The Coverage Rate is then:

\vspace{-6mm}

\[
CR = \frac{\sum_{i=1}^{n} w(s_i) \cdot \mathbb{I}(s_i)}{\sum_{i=1}^{n} w(s_i)},
\]

\vspace{-3mm}

where \( \mathbb{I}(s_i) = 1 \) if subtask \( s_i \) is completed, and 0 otherwise. This metric emphasizes deeper, more complex subtasks, providing a refined measure of agent performance.}

\textls[-30]{\textbf{Logical Consistency~(LC).} It quantifies the operational logic similarity between agents and humans. This metric is motivated by the observation that humans prefer to complete all possible subtasks within an application before switching to another, unless necessary. It is computed as the ratio of the agent's Coherency Score~(CS) to the maximum possible CS:

\vspace{-6mm}

\[
LC = \frac{CS_{agent}}{CS_{max}},
\]

\vspace{-4mm}

where \( CS \) quantifies the coherence of the subtask sequence. For each pair of adjacent subtasks \((s_i, s_{i+1})\) in sequence, \( CS \) increases by 1 if both subtasks belong to the same application. \( CS_{agent} \) is the coherence score calculated from the executing subtask sequence, and \( CS_{max} \) is the maximum possible coherence score calculated among all topological sequences.}

\vspace{-2mm}
\subsection{Evaluation Strategy}
\vspace{-2mm}
\label{sec:4.2}
\textls[-30]{Currently, there is limited discussion on the categorization of capabilities in the virtual agent field. The previous mainstream classifications~\cite{showui,uground} typically divided the capabilities into two simple categories: grounding and planning. However, this simple classification is quite coarse and does not take into account other essential and fine-grained capabilities for agents, such as decision-making and instruction understanding. To address this, we propose 10 fine-grained capabilities, derived from five categories that we consider essential, with each category contributing two capabilities for agents. Specific test tasks are constructed for each capability based on the combination of five complexity dimensions, as shown in Figure~\ref{mainfigure}.}

\textls[-30]{Taking long-range planning capability as an example, we categorize tasks with higher dependency complexity and hierarchy complexity as test tasks for this capability. This is because higher dependency complexity means the task involves more dependencies, requiring stronger planning capability. Meanwhile, higher hierarchy complexity indicates the task has deeper levels, which places higher demands on long-sequence processing capability. Therefore, we select tasks with dependency complexity and hierarchy complexity at the hard level as test tasks for long-sequence reasoning capability. For the test tasks of these 10 capabilities, we engaged professionally trained annotators to filter and construct high-quality test data. The specific definitions and corresponding explanations for the other 9 capabilities can be found in Appendix\textcolor{red}{\hypersetup{linkcolor=red}\hyperref[compl_and_cap]{~\ref{compl_and_cap}}}.}

\vspace{-3mm}
\section{Experiments}
\vspace{-2mm}
\textls[-30]{In this section, we first introduce the experimental setup~(Section~\ref{sec:5.1}). Then, we comprehensively compare the differences in capabilities across various models on OmniBench, along with several key findings~(Section~\ref{sec:5.2}). 
Finally, we provide an in-depth analysis of the reasons behind the poor performance and methods for performance improvement~(Section~\ref{sec:5.3}).}

\vspace{-2mm}
\subsection{Experimental Setup}
\vspace{-2mm}
\label{sec:5.1}

\begin{table*}[h!]
\vspace{-6mm}
\centering
\caption{Performance of models on OmniBench. For each capability, we use the CR metric on test tasks for quantification. Abbreviations adopted: PP for Parallel Planning; LRP for Long Range Planning; CDDK for Cross-Domain Decision-Making; SDK for Sequential Decision-Making; SI for Subtask Identification; DI for Dependency Identification; LSR for Long Sequence Reasoning; LIF for Long Instruction Following; DSK for Domain-Specific Knowledge; CDK for Cross-Domain Knowledge. An asterisk (*) indicates that the agent uses GPT-4o as the planner.}
\resizebox{\textwidth}{!}{%
\begin{tabular}{lcccccccccccc}
 \toprule 
  \multirow{2}{*}{} & \multicolumn{2}{c}{\textbf{\makecell{Overall}}} & \multicolumn{2}{c}{\textbf{\makecell{Planning}}} & \multicolumn{2}{c}{\textbf{\makecell{Decision-making}}} & \multicolumn{2}{c}{\textbf{\makecell{Instruction\\Understanding}}} & \multicolumn{2}{c}{\textbf{\makecell{Long Context}}} & \multicolumn{2}{c}{\textbf{\makecell{Generalist\\Knowledge}}} \\ \cline{2-13}
  & $CR$ & $LC$ & $PP$ & $LRP$ & $CDDK$ & $SDK$ & $SI$ & $DI$ & $LSR$ & $LIF$ & $DSK$ & $CDK$ \\ \midrule
 Human & 80.1 & 92.8 & 80.1 & 76.9 & 91.9 & 93.0 & 69.1 & 72.1 & 79.5 & 66.1 & 89.4 & 71.5 \\ \midrule
 \rowcolor{orange!20}
 \multicolumn{13}{c}{\textit{Open-source Multimodal Large Language Models~(A11Y+Screenshot)}} \\
 Qwen2-VL-7B~\cite{qwen2vl} & 14.8 & 9.0 & 15.5 & 13.5 & 16.5 & 17.8 & 14.1 & 13.8 & 14.7 & 12.4 & 15.8 & 13.8 \\
 InternVL2-8B~\cite{internvl} & 14.2 & 13.0 & 15.0 & 13.5 & 16.2 & 16.9 & 12.1 & 12.9 & 15.8 & 11.7 & 15.8 & 12.4 \\
 InternVL2.5-8B~\cite{internvl} & 17.4 & 18.8 & 18.2 & 16.7 & 19.6 & 21.5 & 16.4 & 15.3 & 15.8 & 15.4 & 19.0 & 16.3 \\ 
  \rowcolor{blue!20}
 \multicolumn{13}{c}{\textit{Closed-source Multimodal Large Language Models~(A11Y+Screenshot)}} \\
Qwen-VL-Max~\cite{qwenvl} & 18.4 & 23.3 & 18.7 & 19.4 & 19.6 & 23.3 & 15.0 & 16.7 & 18.3 & 16.1 & 19.6 & 17.3 \\ 
Gemini-2.0-Flash & 25.9 & \underline{38.0} & 24.8 & 24.6 & 31.5 & \underline{33.2} & 22.5 & 22.5 & 25.7 & 21.9 & 27.8 & \underline{24.8} \\ 
Claude-3.5-Sonnet & \underline{27.6} & 35.0 & \underline{30.5} & \underline{24.7} & \underline{32.0} & 31.3 & \underline{24.5} & \underline{25.0} & \underline{26.6} & \underline{23.5} & \underline{33.4} & 24.5 \\ 
GPT-4o~\cite{gpt} & \textbf{38.7} & \textbf{49.0} & \textbf{38.4} & \textbf{37.8} & \textbf{43.2} & \textbf{49.4} & \textbf{30.6} & \textbf{35.5} & \textbf{42.7} & \textbf{32.2} & \textbf{43.2} & \textbf{34.2} \\ 
\midrule
  \rowcolor{green!20} 
 \multicolumn{13}{c}{\textit{Visual Digital Agents~(Screenshot)}} \\
 Aguvis-7B~\cite{aguvis} & 22.9 & {27.1} & 21.2 & 23.5 & 25.5 & 28.1 & 20.2 & 20.0 & 22.8 & 20.1 & 26.3 & 21.6 \\
 OS-Atlas-Pro-4B~\cite{os-atlas} & {19.1} & 23.9 & 20.6 & 17.6 & 22.9 & 23.6 & 15.0 & 17.7 & 18.7 & 15.9 & 22.0 & 16.8 \\
 ShowUI-2B\textsuperscript{*}~\cite{showui} & 23.2 & 24.6 & 23.2 & 23.1 & 26.3 & 26.6 & 21.5 & 20.3 & 24.7 & 20.4 & 24.8 & 20.7 \\
 OS-Atlas-Base-4B\textsuperscript{*}~\cite{os-atlas} & 22.2 & 23.8 & 23.2 & 21.9 & 26.2 & 25.6 & 19.4 & 19.5 & 23.5 & 20.0 & 23.4 & 19.3 \\
 UGround-V1-7B\textsuperscript{*}~\cite{uground} & 25.0 & 26.3 & \underline{25.7} & {25.1} & 30.6 & {31.4} & 21.5 & 21.3 & {24.8} & 21.3 & 27.2 & {21.5} \\
  \rowcolor{purple!20}
 \multicolumn{13}{c}{\textit{Supervised Fine-Tuning Agents~(Screenshot)}} \\
  Omni-OS-Atlas-Base-4B~(Ours) & \underline{29.7} & \underline{30.1} & 24.2 & \textbf{33.0} & \underline{34.9} & \underline{35.3} & \textbf{28.7} & \underline{24.2} & \underline{27.9} & \underline{26.5} & \underline{33.8} & \textbf{28.2} \\
 Omni-UGround-V1-7B~(Ours)  & \textbf{34.4} & \textbf{37.4} & \textbf{33.2} & \underline{31.3} & \textbf{43.1} & \textbf{42.4} & \underline{21.9} & \textbf{35.0} & \textbf{40.3} & \textbf{31.7} & \textbf{36.7} & \underline{27.6} \\
 
\bottomrule
\end{tabular}
}
\label{maintable}
\vspace{-7mm}
\end{table*}

\textls[-30]{\textbf{Settings.} We evaluate various models including MLLMs and Virtual Agents on OmniBench. For all virtual agents, we use the default prompt provided by each agent for inference, if available. If models do not provide prompts for agent tasks, we use a unified prompt designed by us. We also report results trained on OmniBench data for some selected models. All experiments are conducted with NVIDIA A100 80G GPUs.}

\textls[-30]{\textbf{Baselines.} We conduct a comprehensive evaluation of the four types of models as shown in Table~\ref{maintable}.
The specific details about the baselines can be found in Appendix\textcolor{red}{\hypersetup{linkcolor=red}\hyperref[baseline]{~\ref{baseline}}}.}

\vspace{-2mm}
\subsection{Main Results}
\begin{figure}[htbp]
    \centering
    \vspace{-9mm}
    \includegraphics[width=\columnwidth]{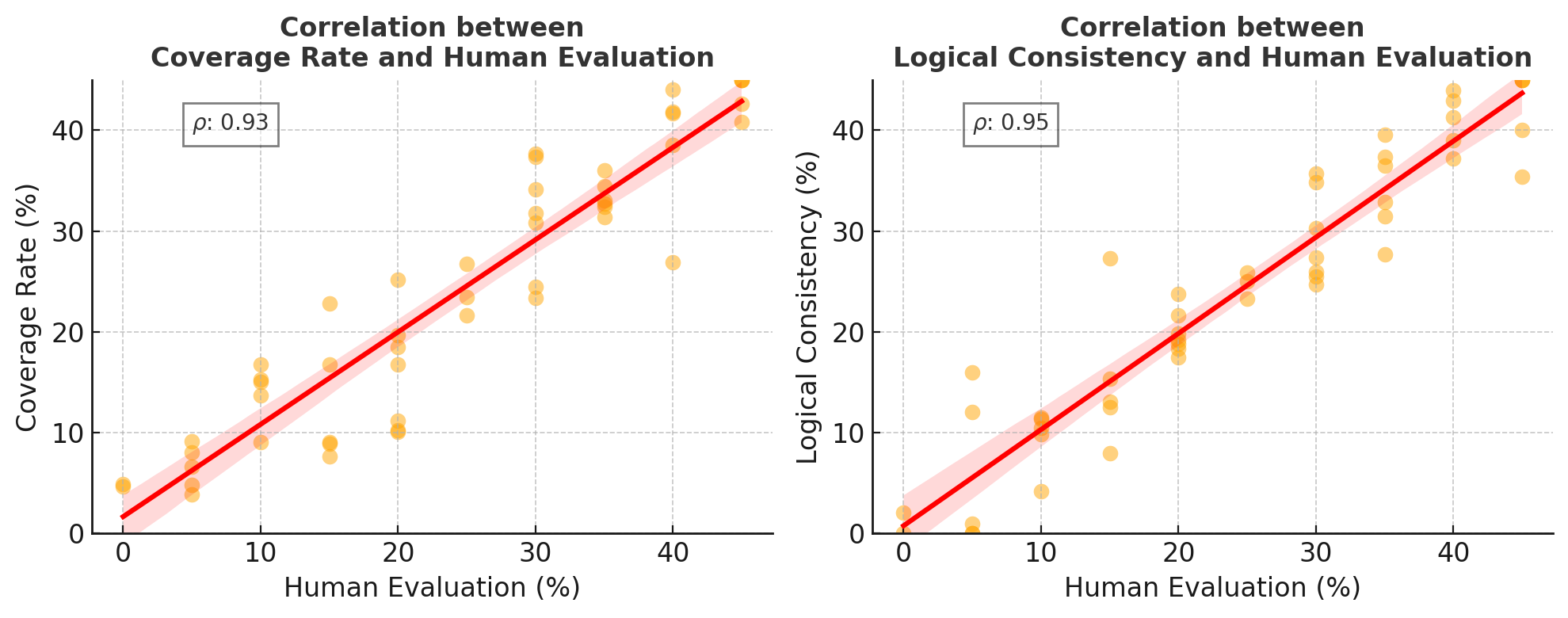}
    \vspace{-9mm}
    \caption{Correlation between Coverage Rate and Logical Consistency with Human Evaluation.}
    \label{cohere}
\end{figure}

\vspace{-2mm}
\label{sec:5.2}
\textls[-30]{\noindent\textbf{Alignment Between OmniEval and Human Evaluation.} 
Before delving deeper into the concrete evaluation results, we first compare the alignment between \mbox{OmniEval} and human evaluation for agent tasks. Specifically, we randomly sampled 50 trajectories from all models to calculate the correlation between OmniEval and human evaluation. Each trajectory was scored by two specially trained annotators, who referenced the task instructions to assign task completion scores and logical alignment scores from \{0\%, 10\%, ..., 90\%, 100\%\}. The final human evaluation score was determined by averaging the scores given by the two annotators. In Figure~\ref{cohere}, we present the Pearson correlation between OmniEval and human evaluation. The results indicate a strong correlation between OmniEval and human evaluation.}

\begin{figure}[htbp]
    \centering
    \vspace{-5mm}
    \includegraphics[width=\columnwidth]{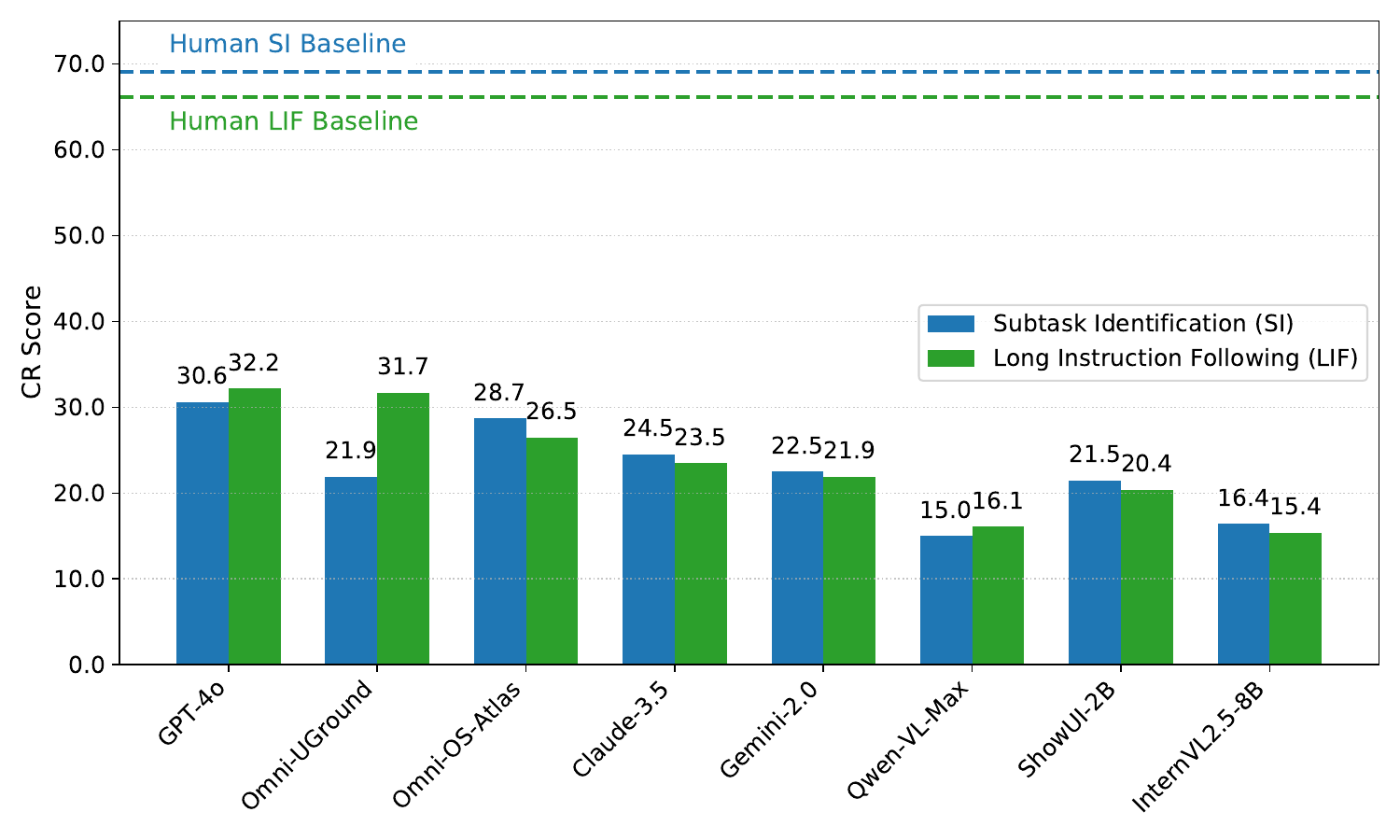}
    \vspace{-10mm}
    \caption{Comparison of model performance on Subtask Identification (SI) and Long Instruction Following (LIF).}
    \label{bound}
    \vspace{-7mm}
\end{figure}

\textls[-30]{\noindent\textbf{Capability Boundaries of Mainstream Agents.} As detailed in Table~\ref{maintable}, although advanced agents such as GPT-4o and our supervised fine-tuning models (e.g., Omni-UGround) demonstrate strong performance in overall metrics and in capabilities like planning and decision-making, clear limitations remain in Subtask Identification (SI) and Long Instruction Following (LIF). Specifically, even the strongest models only achieve 30.6 (GPT-4o) and 21.9 (Omni-UGround) in SI, and 32.2 and 31.7 respectively in LIF, which are significantly lower than the human baselines of 69.1 and 66.1, as shown in Figure~\ref{bound}. These results highlight a persistent difficulty in decomposing complex instructions and maintaining coherence over extended task flows. Compared to their performance in other capabilities, the results suggest that instruction understanding in long and semantically complex contexts remains a key bottleneck for current agents. Future improvements in agent performance will likely depend on more robust handling of multi-step semantics and long-context alignment.}

\newcommand{\droptext}[1]{{\scriptsize\textcolor{red}{#1}}}
\newcommand{\uptext}[1]{{\scriptsize\textcolor{blue}{#1}}}

\begin{table*}[h]
    \centering
    \vspace{-6mm}
    \caption{We compare the performance of different virtual agents on tasks with varying complexity levels. Medium and Hard levels show performance drops compared to the previous level, with downward arrows indicating the magnitude of decline.}
    \setlength{\tabcolsep}{4pt}
    \resizebox{\textwidth}{!}{%
    \begin{tabular}{lcccccccccccccccc}
        \toprule
        \multirow{2}{*}{\textbf{Agents}} & \multicolumn{3}{c}{\textbf{Dependency Comp.} ↑} & \multicolumn{3}{c}{\textbf{Branch Comp.} ↑} & \multicolumn{3}{c}{\textbf{Instruction Comp.} ↑} & \multicolumn{3}{c}{\textbf{Knowledge Comp.} ↑} & \multicolumn{3}{c}{\textbf{Hierarchy Comp.} ↑} \\
        & Easy & Medium & Hard & Easy & Medium & Hard & Easy & Medium & Hard & Easy & Medium & Hard & Easy & Medium & Hard \\
        \midrule
Aguvis-7B~\cite{aguvis} & 32.8 & 27.6 \droptext{$\downarrow$5.2} & 24.3 \droptext{$\downarrow$3.3} & 41.2 & 36.8 \droptext{$\downarrow$4.4} & 30.6 \droptext{$\downarrow$6.2} & 49.5 & 36.9 \droptext{$\downarrow$12.6} & 25.3 \droptext{$\downarrow$11.6} & 38.4 & 32.5 \droptext{$\downarrow$5.9} & 27.6 \droptext{$\downarrow$4.9} & 37.9 & 33.6 \droptext{$\downarrow$4.3} & 29.7 \droptext{$\downarrow$3.9} \\
OS-Atlas-Pro-7B~\cite{os-atlas} & 32.3 & 26.8 \droptext{$\downarrow$5.5} & 23.7 \droptext{$\downarrow$3.1} & 39.1 & 31.0 \droptext{$\downarrow$8.1} & 25.4 \droptext{$\downarrow$5.6} & 44.3 & 34.8 \droptext{$\downarrow$9.5} & 21.8 \droptext{$\downarrow$13.0} & 33.9 & 28.4 \droptext{$\downarrow$5.5} & 24.3 \droptext{$\downarrow$4.1} & 34.5 & 28.1 \droptext{$\downarrow$6.4} & 25.6 \droptext{$\downarrow$2.5} \\
ShowUI-2B\textsuperscript{*}~\cite{showui} & 34.0 & 28.3 \droptext{$\downarrow$5.7} & 25.6 \droptext{$\downarrow$2.7} & 41.3 & 32.7 \droptext{$\downarrow$8.6} & 28.2 \droptext{$\downarrow$4.5} & 45.9 & 36.6 \droptext{$\downarrow$9.3} & 25.4 \droptext{$\downarrow$11.2} & 37.8 & 32.6 \droptext{$\downarrow$5.2} & 27.4 \droptext{$\downarrow$5.2} & 37.6 & 32.0 \droptext{$\downarrow$5.6} & 28.1 \droptext{$\downarrow$3.9} \\
OS-Atlas-Base-4B\textsuperscript{*}~\cite{os-atlas} & 32.7 & 29.1 \droptext{$\downarrow$3.6} & 24.9 \droptext{$\downarrow$4.2} & 35.2 & 32.4 \droptext{$\downarrow$2.8} & 27.6 \droptext{$\downarrow$4.8} & 48.2 & 37.5 \droptext{$\downarrow$10.7} & 26.7 \droptext{$\downarrow$10.8} & 39.1 & 34.2 \droptext{$\downarrow$4.9} & 28.9 \droptext{$\downarrow$5.3} & 43.1 & 38.4 \droptext{$\downarrow$4.7} & 33.2 \droptext{$\downarrow$5.2} \\
UGround-7B\textsuperscript{*}~\cite{uground} & 34.1 & 30.0 \droptext{$\downarrow$4.1} & 27.1 \droptext{$\downarrow$2.9} & 44.6 & 38.3 \droptext{$\downarrow$6.3} & 32.4 \droptext{$\downarrow$5.9} & 53.0 & 39.4 \droptext{$\downarrow$13.6} & 27.2 \droptext{$\downarrow$12.2} & 42.3 & 36.4 \droptext{$\downarrow$5.9} & 32.6 \droptext{$\downarrow$3.8} & 35.7 & 28.8 \droptext{$\downarrow$6.9} & 25.5 \droptext{$\downarrow$3.3} \\
        \bottomrule
    \end{tabular}%
    }
    \label{complexity}
    \vspace{-7mm}
\end{table*}

\begin{figure}[ht!]
\vspace{-4mm}
\includegraphics[width=\linewidth]{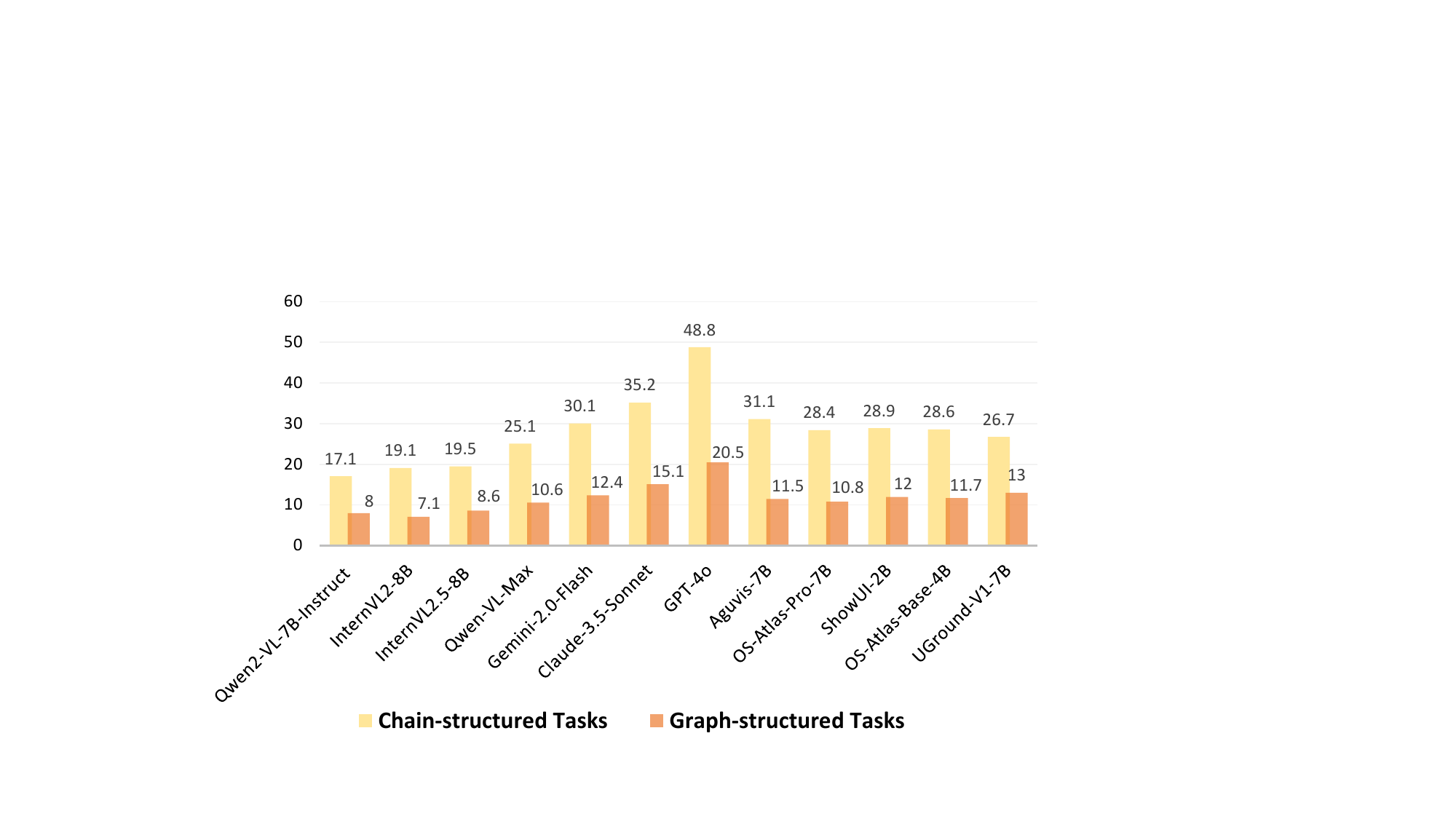}
\vspace{-8mm}
\centering\caption{Performance comparison of various models on chain-structured tasks and graph-structured tasks.}
\label{chainvsgraph}
\vspace{-3mm}
\end{figure}

\textls[-30]{\textbf{Challenges in Handling Graph-structured Tasks.} We compared the performance differences of each model on chain-structured tasks and graph-structured tasks.
To eliminate the influence of other factors, we utilized OmniBench's controllable task synthesis mechanism to construct a set of chain-structured and graph-structured tasks, each with the same number of nodes and edges. All tasks belong to the same knowledge domain and share the same level of knowledge complexity. As shown in Figure~\ref{chainvsgraph}, GPT-4o (with A11Y), the most advanced agent, achieves only 20.5\% accuracy on graph-structured tasks, far below the human performance at 80.1\%. This phenomenon can be attributed to the fact that most existing agents are predominantly fine-tuned on chain-structured tasks, which may result in their tendency to interpret graph-structured tasks as linear. Such misinterpretation can significantly impair the agents' capability to accurately identify the dependency relationships between subtasks, ultimately leading to task execution failures.}

\begin{table}[htbp]
\vspace{-6mm}
\centering
\caption{Evaluation results on AndroidControl benchmark. \textbf{Bold} values indicate the best performance across all baselines.}
\resizebox{\linewidth}{!}{
\begin{tabular}{l|ccc|ccc}
\hline
\multirow{2}{*}{\textbf{Models}} & 
\multicolumn{3}{c|}{\textbf{AndroidControl-Low}} & 
\multicolumn{3}{c}{\textbf{AndroidControl-High}} \\
& \textbf{Type} & \textbf{Grounding} & \textbf{SR} 
& \textbf{Type} & \textbf{Grounding} & \textbf{SR} \\
\hline
InternVL-2-4B~\cite{internvl}       & 90.94 & 84.05 & 80.10 & 84.09 & 72.73 & 66.72 \\
Qwen2-VL-7B~\cite{qwen2vl}         & 91.94 & 86.50 & 82.56 & 83.83 & 77.68 & 69.72 \\
SeeClick~\cite{seeclick}            & 93.00 & 73.42 & 75.00 & 82.94 & 62.87 & 59.11 \\
OS-Atlas-4B~\cite{os-atlas}         & 91.92 & 83.76 & 80.64 & 84.69 & 73.79 & 67.54 \\
UGround-7B-V1~\cite{uground}       & 92.15 & 87.17 & 83.29 & 84.72 & 78.85 & 70.31 \\
Omni-OS-Atlas-4B~(Ours)    & \textbf{92.49} & 83.51 & 81.38 & 84.86 & 73.81 & 67.71 \\
Omni-UGround-7B-V1~(Ours)  & 92.37 & \textbf{87.24} & \textbf{83.57} & \textbf{84.89} & \textbf{78.97} & \textbf{70.83} \\
\hline
\end{tabular}
}
\label{android}
\end{table}

\begin{table}[htbp]
\vspace{-9mm}
\centering
\caption{Evaluation results on OmniAct benchmark. \textbf{Bold} values indicate the best performance across all baselines.}
\resizebox{\linewidth}{!}{
\begin{tabular}{l|ccc|ccc}
\hline
\multirow{2}{*}{\textbf{Models}} & 
\multicolumn{3}{c|}{\textbf{OmniAct-Web}} & 
\multicolumn{3}{c}{\textbf{OmniAct-Desktop}} \\
& \textbf{Type} & \textbf{Grounding} & \textbf{SR} 
& \textbf{Type} & \textbf{Grounding} & \textbf{SR} \\
\hline
InternVL-2-4B~\cite{internvl}       & 47.51 & 51.34 & 24.39 & 67.00 & 44.47 & 29.80 \\
Qwen2-VL-7B~\cite{qwen2vl}         & 89.22 & 85.94 & 78.58 & 96.27 & 94.52 & 91.77 \\
SeeClick~\cite{seeclick}            & 86.98 & 75.48 & 68.59 & 96.79 & 70.22 & 72.59 \\
OS-Atlas-4B~\cite{os-atlas}         & 88.56 & 82.00 & 73.91 & 96.51 & 85.53 & 84.78 \\
UGround-7B-V1~\cite{uground}       & 90.16 & 86.98 & 79.85 & 97.13 & 94.79 & 91.89 \\
Omni-OS-Atlas-4B~(Ours)    & 89.96 & 82.74 & 74.62 & 97.64 & 86.37 & 85.53 \\
Omni-UGround-7B-V1~(Ours)  & \textbf{91.24} & \textbf{87.35} & \textbf{80.24} & \textbf{97.93} & \textbf{95.21} & \textbf{92.10} \\
\hline
\end{tabular}
}
\vspace{-5mm}
\label{omniact}
\end{table}

\vspace{-2mm}
\subsection{In-Depth Analysis}
\vspace{-2mm}
\label{sec:5.3}

\textls[-30]{\noindent\textbf{Performance Differences Across Complexity Levels.}
As shown in Table~\ref{complexity}, we analyze agent performance across tasks grouped by complexity levels: Easy, Medium, and Hard. Unsurprisingly, all agents exhibit significant performance drops as task complexity increases, with an average decrease of 6.19 points. This trend is consistent across all five dimensions. Such systematic degradation in harder cases confirms OmniBench's effectiveness in scaling task difficulty. Furthermore, performance on hard tasks may serve as a more accurate indicator of an agent's expert capabilities than average scores, revealing the upper bounds of its potential.}

\textls[-30]{\textbf{Effectiveness of Graph-Structured Task Trajectories.}
We follow the training details provided in the OS-Atlas paper and adopt the same experimental setup to train our backbone models: OS-Atlas-4B and UGround-7B-V1. As shown in Table~\ref{android} and Table~\ref{omniact}, while agents such as OS-Atlas and UGround, which are pretrained on GUI grounding tasks, exhibit strong GUI understanding capabilities, their limited planning capability hinders their performance in complex action reasoning. 
In contrast, the high-quality multi-step navigation dataset synthesized by OmniBench significantly enhances the model’s capability to make decisions regarding action types, thereby improving the success rate in GUI navigation. 
Specifically, Omni-OS-Atlas-4B achieves an average success rate improvement of 0.46 points on AndroidControl and 0.73 points on OmniAct, while Omni-UGround-7B-V1 achieves improvements of 0.4 points on AndroidControl and 0.3 points on OmniAct.}

\textls[-30]{\textbf{Sensitivity to Expression Order in Task Instructions.} 
We define the impact of textual order on the model as its instruction sensitivity, conducting experiments with standard deviation as the metric. We construct 10 specially designed test tasks, each associated with three task instructions that are semantically identical (based on the same task graph) but differ in textual order. As shown in Table~\ref{tab:sensitivity}, the original MLLMs tend to be less sensitive to instruction variations, but perform poorly overall. Though fine-tuning them on navigation tasks enhances the performance, it also compromises the models' robustness to instructions. OS-Atlas-Pro and Aguvis exhibit significantly higher sensitivity, with an average increase of 8.21 points. Moreover, after incorporating graph-structured task samples from OmniBench into fine-tuning, the models' performance is further improved while largely preserving their robustness. Omni-OS-Atlas and Omni-Aguvis exhibit reduced sensitivity, with an average reduction of 7.91 points. This indicates that the diverse and structured task trajectories from OmniBench can help models better recognize complex dependencies embedded in task instructions, improving their overall stability and performance.}

\begin{table}[h]
\centering
\vspace{-5mm}
\caption{Average sensitivity across different models.}
\resizebox{\linewidth}{!}{%
\begin{tabular}{lcc}
\toprule
\textbf{Models} & \textbf{Backbone} & \textbf{Avg. Sensitivity} ↓ \\
\midrule
Human & - & 1.95 \\
\midrule
InternVL2-4B~\cite{internvl} & InternVL2-4B & 2.97 \\
OS-Atlas-Pro~\cite{os-atlas} & InternVL2-4B & 9.07 \uptext{$\uparrow$6.1} \\
Omni-OS-Atlas~(Ours) & InternVL2-4B & 3.49 \droptext{$\downarrow$5.58} \\
\midrule
Qwen2-VL-7B~\cite{qwen2vl} & Qwen2-VL-7B & 2.58 \\
Aguvis~\cite{aguvis} & Qwen2-VL-7B & 12.9 \uptext{$\uparrow$10.32} \\
Omni-Aguvis~(Ours) & Qwen2-VL-7B & 2.67 \droptext{$\downarrow$10.23} \\
\bottomrule
\end{tabular}%
}
\vspace{-6mm}
\label{tab:sensitivity}
\end{table}

\textls[-30]{\textbf{The Effect of Task Intent on Planning.}
We design two experiments to explore the applicability of task intent to both open-source and closed-source models. 
\textbf{1) For open-source models}, we conduct a comparative experiment using two separate datasets to fine-tune OS-Atlas-Base-4B and UGround-V1-7B. One dataset includes task intent, while the other does not. As shown in Table~\ref{tab:intent}, incorporating task intent in the training data significantly improves the model's planning performance on OmniBench. Specifically, OS-Atlas-Base-4B improves its overall planning score from 28.6 to 30.3, an increase of 1.7 points, while UGround-V1-7B improves from 32.3 to 33.5, gaining 1.2 points. 
\textbf{2) For closed-source models}, we use Qwen-VL-Max, Gemini-2.0-Flash, Claude-3.5-Sonnet, and GPT-4o as planners, with UGround-V1-7B serving as the grounding models. As shown in Table~\ref{tab:intent}, all closed-source models exhibit improved planning performance with the inclusion of task intent. GPT-4o shows the most significant improvement, with its overall score rising from 25.4 to 34.3, a gain of 8.9 points. Claude-3.5-Sonnet increases by 5.4 points, Gemini-2.0-Flash by 4.9 points, and Qwen-VL-Max by 2.6 points. This indicates that closed-source models can enhance their planning through this plug-and-play approach.}

\begin{table}[h]
\centering
\vspace{-5mm}
\caption{The effect of task intent on the planning of open-source versus closed-source models.}
\vspace{1mm}
\resizebox{\linewidth}{!}{%
\begin{tabular}{lccc}
\toprule
\textbf{Models} & \textbf{Parallel Planning} ↑ & \textbf{Long-Range Planning} ↑ & \textbf{Overall} ↑ \\
\midrule
\rowcolor{orange!20}
\multicolumn{4}{c}{\textit{Open-source Multimodal Large Language Models}} \\
Omni-OS-Atlas-Base-4B & 24.2 & 33.0 & 28.6 \\
\ \ \ \ + intent tuning & 25.7 \uptext{$\uparrow$1.5} & 34.9 \uptext{$\uparrow$1.9} & 30.3 \uptext{$\uparrow$1.7} \\
Omni-UGround-V1-7B & 33.2 & 31.3 & 32.3 \\
\ \ \ \ + intent tuning & 34.4 \uptext{$\uparrow$1.2} & 32.6 \uptext{$\uparrow$1.3} & 33.5 \uptext{$\uparrow$1.2} \\
\midrule
\rowcolor{blue!20}
\multicolumn{4}{c}{\textit{Closed-source Multimodal Large Language Models}} \\
Qwen-VL-Max & 21.9 & 20.8 & 21.4 \\
\ \ \ \ + intent prompt & 24.5 \uptext{$\uparrow$2.6} & 23.5 \uptext{$\uparrow$2.7} & 24.0 \uptext{$\uparrow$2.6} \\
Gemini-2.0-Flash & 23.1 & 22.7 & 22.9 \\
\ \ \ \ + intent prompt & 28.9 \uptext{$\uparrow$5.8} & 26.7 \uptext{$\uparrow$4.0} & 27.8 \uptext{$\uparrow$4.9} \\
Claude-3.5-Sonnet & 24.2 & 23.7 & 24.0 \\
\ \ \ \ + intent prompt & 30.6 \uptext{$\uparrow$6.4} & 28.1 \uptext{$\uparrow$4.4} & 29.4 \uptext{$\uparrow$5.4} \\
GPT-4o & 25.7 & 25.1 & 25.4 \\
\ \ \ \ + intent prompt & 32.9 \uptext{$\uparrow$7.2} & 35.7 \uptext{$\uparrow$10.6} & 34.3 \uptext{$\uparrow$8.9} \\
\bottomrule
\end{tabular}%
}
\vspace{-3mm}
\label{tab:intent}
\end{table}

\begin{figure}[h!]
\vspace{-2mm}
\includegraphics[width=\linewidth]{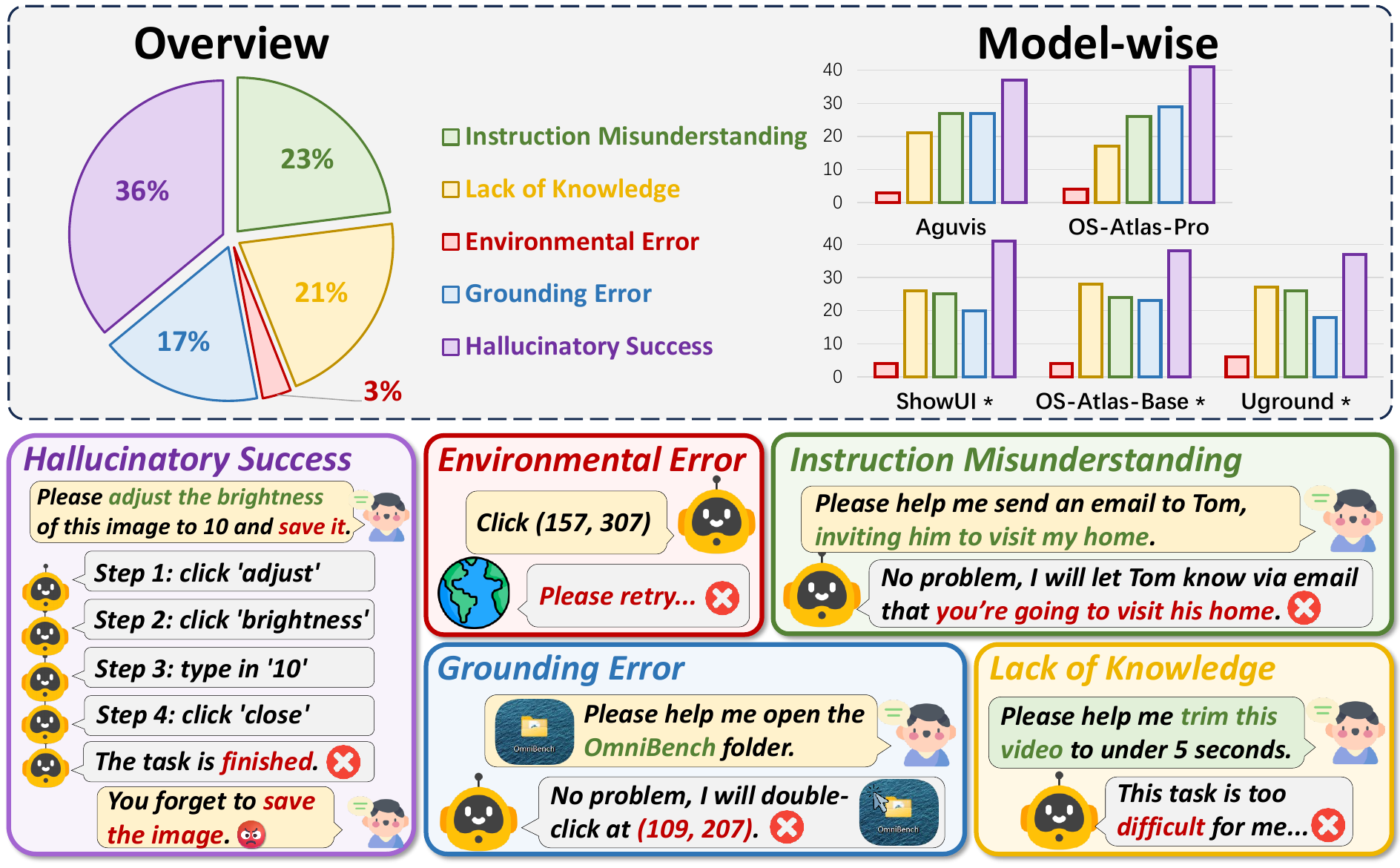}
\vspace{-8mm}
\centering\caption{Distribution of five major errors in 100 failure instances for each model. An asterisk (*) indicates that the agent uses GPT-4o as the planner.}
\label{error}
\end{figure}

\textbf{Failure Analysis.} 
In this section, we delve into the analysis of errors encountered during the OmniBench evaluation. This analysis aims not only to identify the current shortcomings of the agents but also to inform future improvements in their design and training. We carefully examine 100 randomly sampled error instances for each model from the OmniBench evaluation. These instances are analyzed by expert annotators who identify the root causes of mispredictions based on their knowledge. Specifically, there are five types of errors: 
\textbf{1)~Instruction Understanding.} We observe that 23\% of the failures are due to the agent's misunderstanding of the instructions. For example, it overlooks the final save file operation requested in the image editing instruction. 
\textbf{2)~Lack of Knowledge.} We find that 21\% of the failures are caused by the agent's lack of knowledge about the target application, such as being unfamiliar with how to create a reference list in Zotero. 
\textbf{3)~Environmental Error.} We observe that 3\% of the failures result from environmental interference, such as network delays. 
\textbf{4)~Grounding Error.} We find that 17\% of the failures are due to the model's lack of grounding ability, meaning the agent knows the target to click next but locates it in the wrong position. 
\textbf{5)~Hallucinatory Success.} Finally, 36\% of the failures occur when the agent incorrectly assumes the task is complete, which may stem from its weak contextual memory capabilities.
The distribution of these errors is shown in Figure~\ref{error}.

\vspace{-3mm}
\section{Conclusion}
\vspace{-2mm}
In conclusion, we introduced \textbf{\mbox{OmniBench}}, a graph-based benchmark that addresses the limitations of existing evaluation frameworks by enabling controllable task complexity through automated subtask composition. Along with \textbf{\mbox{OmniEval}}, a multidimensional evaluation framework, we evaluate virtual agents across 10 capabilities. Our results show that training on this data improves agent generalization, and our evaluations provide valuable insights into the strengths and areas for improvement in MLLM-based virtual agents.

\textbf{Acknowledgement.} This work was supported by the NSFC (62272411), the Fundamental Research Funds for the Central Universities (226-2025-00017), the Key R\&D Projects in Zhejiang Province (No. 2024C01106, 2025C01030), Ningbo Yongjiang Talent Introduction Programme (2024A-401-G), the Zhejiang NSF (LRG25F020001),  Ant Group.

\vspace{-3mm}
\section*{Impact Statement}

This paper presents work whose goal is to advance the field of Machine Learning. There are many potential societal consequences of our work, none which we feel must be specifically highlighted here.

\bibliography{example_paper}

\begin{thebibliography}{46}
\providecommand{\natexlab}[1]{#1}
\providecommand{\url}[1]{\texttt{#1}}
\expandafter\ifx\csname urlstyle\endcsname\relax
  \providecommand{\doi}[1]{doi: #1}\else
  \providecommand{\doi}{doi: \begingroup \urlstyle{rm}\Url}\fi

\bibitem[Bai et~al.(2023)Bai, Bai, Yang, Wang, Tan, Wang, Lin, Zhou, and Zhou]{qwenvl}
Bai, J., Bai, S., Yang, S., Wang, S., Tan, S., Wang, P., Lin, J., Zhou, C., and Zhou, J.
\newblock Qwen-vl: A versatile vision-language model for understanding, localization, text reading, and beyond.
\newblock \emph{arXiv preprint arXiv:2308.12966}, 2023.

\bibitem[Burns et~al.(2022)Burns, Arsan, Agrawal, Kumar, Saenko, and Plummer]{motif}
Burns, A., Arsan, D., Agrawal, S., Kumar, R., Saenko, K., and Plummer, B.~A.
\newblock A dataset for interactive vision-language navigation with unknown command feasibility.
\newblock In \emph{European Conference on Computer Vision}, pp.\  312--328. Springer, 2022.

\bibitem[Cao et~al.(2024)Cao, Lei, Wu, Chen, Fu, Gao, Xiong, Zhang, Mao, Hu, et~al.]{spider2-v}
Cao, R., Lei, F., Wu, H., Chen, J., Fu, Y., Gao, H., Xiong, X., Zhang, H., Mao, Y., Hu, W., et~al.
\newblock Spider2-v: How far are multimodal agents from automating data science and engineering workflows?
\newblock \emph{arXiv preprint arXiv:2407.10956}, 2024.

\bibitem[Chen et~al.(2024)Chen, Wu, Wang, Su, Chen, Xing, Zhong, Zhang, Zhu, Lu, et~al.]{internvl}
Chen, Z., Wu, J., Wang, W., Su, W., Chen, G., Xing, S., Zhong, M., Zhang, Q., Zhu, X., Lu, L., et~al.
\newblock Internvl: Scaling up vision foundation models and aligning for generic visual-linguistic tasks.
\newblock In \emph{Proceedings of the IEEE/CVF Conference on Computer Vision and Pattern Recognition}, pp.\  24185--24198, 2024.

\bibitem[Cheng et~al.(2024)Cheng, Sun, Chu, Xu, Li, Zhang, and Wu]{seeclick}
Cheng, K., Sun, Q., Chu, Y., Xu, F., Li, Y., Zhang, J., and Wu, Z.
\newblock Seeclick: Harnessing gui grounding for advanced visual gui agents.
\newblock \emph{arXiv preprint arXiv:2401.10935}, 2024.

\bibitem[Deng et~al.(2024)Deng, Gu, Zheng, Chen, Stevens, Wang, Sun, and Su]{mind2web}
Deng, X., Gu, Y., Zheng, B., Chen, S., Stevens, S., Wang, B., Sun, H., and Su, Y.
\newblock Mind2web: Towards a generalist agent for the web.
\newblock \emph{Advances in Neural Information Processing Systems}, 36, 2024.

\bibitem[Fei et~al.(2024{\natexlab{a}})Fei, Wu, Ji, Zhang, and Chua]{fei2024dysen}
Fei, H., Wu, S., Ji, W., Zhang, H., and Chua, T.-S.
\newblock Dysen-vdm: Empowering dynamics-aware text-to-video diffusion with llms.
\newblock In \emph{Proceedings of the IEEE/CVF Conference on Computer Vision and Pattern Recognition}, pp.\  7641--7653, 2024{\natexlab{a}}.

\bibitem[Fei et~al.(2024{\natexlab{b}})Fei, Wu, Ji, Zhang, Zhang, Lee, and Hsu]{fei2024video}
Fei, H., Wu, S., Ji, W., Zhang, H., Zhang, M., Lee, M.-L., and Hsu, W.
\newblock Video-of-thought: Step-by-step video reasoning from perception to cognition.
\newblock In \emph{Proceedings of the International Conference on Machine Learning}, 2024{\natexlab{b}}.

\bibitem[Fei et~al.(2024{\natexlab{c}})Fei, Wu, Zhang, Chua, and Yan]{fei2024vitron}
Fei, H., Wu, S., Zhang, H., Chua, T.-S., and Yan, S.
\newblock Vitron: A unified pixel-level vision llm for understanding, generating, segmenting, editing.
\newblock 2024{\natexlab{c}}.

\bibitem[Fei et~al.(2024{\natexlab{d}})Fei, Wu, Zhang, Zhang, Chua, and Yan]{fei2024enhancing}
Fei, H., Wu, S., Zhang, M., Zhang, M., Chua, T.-S., and Yan, S.
\newblock Enhancing video-language representations with structural spatio-temporal alignment.
\newblock \emph{IEEE Transactions on Pattern Analysis and Machine Intelligence}, 2024{\natexlab{d}}.

\bibitem[Gao et~al.(2024{\natexlab{a}})Gao, Bu, Miao, Wu, Li, Li, Tang, Wu, Zhuang, and Wang]{gva}
Gao, M., Bu, W., Miao, B., Wu, Y., Li, Y., Li, J., Tang, S., Wu, Q., Zhuang, Y., and Wang, M.
\newblock Generalist virtual agents: A survey on autonomous agents across digital platforms.
\newblock \emph{arXiv preprint arXiv:2411.10943}, 2024{\natexlab{a}}.

\bibitem[Gao et~al.(2024{\natexlab{b}})Gao, Li, Fei, Pang, Ji, Wang, Lv, Zhang, Tang, and Zhuang]{minghegao_mm1}
Gao, M., Li, J., Fei, H., Pang, L., Ji, W., Wang, G., Lv, Z., Zhang, W., Tang, S., and Zhuang, Y.
\newblock De-fine: Decomposing and refining visual programs with auto-feedback.
\newblock In \emph{Proceedings of the 32nd ACM International Conference on Multimedia}, MM '24, pp.\  7649–7657, New York, NY, USA, 2024{\natexlab{b}}. Association for Computing Machinery.
\newblock ISBN 9798400706868.
\newblock \doi{10.1145/3664647.3681082}.
\newblock URL \url{https://doi.org/10.1145/3664647.3681082}.

\bibitem[Gao et~al.(2025)Gao, Liu, Yue, Wu, Chen, Li, Tang, Wu, Chua, and Zhuang]{gao2025benchmarkingmultimodalcotreward}
Gao, M., Liu, X., Yue, Z., Wu, Y., Chen, S., Li, J., Tang, S., Wu, F., Chua, T.-S., and Zhuang, Y.
\newblock Benchmarking multimodal cot reward model stepwise by visual program, 2025.
\newblock URL \url{https://arxiv.org/abs/2504.06606}.

\bibitem[Ge et~al.(2024)Ge, Li, Pang, Gao, Pan, Lin, Fei, Zhang, Tang, and Zhuang]{iris}
Ge, Z., Li, J., Pang, X., Gao, M., Pan, K., Lin, W., Fei, H., Zhang, W., Tang, S., and Zhuang, Y.
\newblock Iris: Breaking gui complexity with adaptive focus and self-refining.
\newblock \emph{arXiv preprint arXiv:2412.10342}, 2024.

\bibitem[Gou et~al.(2024)Gou, Wang, Zheng, Xie, Chang, Shu, Sun, and Su]{uground}
Gou, B., Wang, R., Zheng, B., Xie, Y., Chang, C., Shu, Y., Sun, H., and Su, Y.
\newblock Navigating the digital world as humans do: Universal visual grounding for gui agents.
\newblock \emph{arXiv preprint arXiv:2410.05243}, 2024.

\bibitem[Hong et~al.(2024)Hong, Wang, Lv, Xu, Yu, Ji, Wang, Wang, Dong, Ding, et~al.]{cogagent}
Hong, W., Wang, W., Lv, Q., Xu, J., Yu, W., Ji, J., Wang, Y., Wang, Z., Dong, Y., Ding, M., et~al.
\newblock Cogagent: A visual language model for gui agents.
\newblock In \emph{Proceedings of the IEEE/CVF Conference on Computer Vision and Pattern Recognition}, pp.\  14281--14290, 2024.

\bibitem[Hu et~al.(2024)Hu, Ouyang, Gao, and Shou]{dawn}
Hu, S., Ouyang, M., Gao, D., and Shou, M.~Z.
\newblock The dawn of gui agent: A preliminary case study with claude 3.5 computer use.
\newblock \emph{arXiv preprint arXiv:2411.10323}, 2024.

\bibitem[Hurst et~al.(2024)Hurst, Lerer, Goucher, Perelman, Ramesh, Clark, Ostrow, Welihinda, Hayes, Radford, et~al.]{gpt}
Hurst, A., Lerer, A., Goucher, A.~P., Perelman, A., Ramesh, A., Clark, A., Ostrow, A., Welihinda, A., Hayes, A., Radford, A., et~al.
\newblock Gpt-4o system card.
\newblock \emph{arXiv preprint arXiv:2410.21276}, 2024.

\bibitem[Kapoor et~al.(2025)Kapoor, Butala, Russak, Koh, Kamble, AlShikh, and Salakhutdinov]{omniact}
Kapoor, R., Butala, Y.~P., Russak, M., Koh, J.~Y., Kamble, K., AlShikh, W., and Salakhutdinov, R.
\newblock Omniact: A dataset and benchmark for enabling multimodal generalist autonomous agents for desktop and web.
\newblock In \emph{European Conference on Computer Vision}, pp.\  161--178. Springer, 2025.

\bibitem[Koh et~al.(2024)Koh, Lo, Jang, Duvvur, Lim, Huang, Neubig, Zhou, Salakhutdinov, and Fried]{visualwebarena}
Koh, J.~Y., Lo, R., Jang, L., Duvvur, V., Lim, M.~C., Huang, P.-Y., Neubig, G., Zhou, S., Salakhutdinov, R., and Fried, D.
\newblock Visualwebarena: Evaluating multimodal agents on realistic visual web tasks.
\newblock \emph{arXiv preprint arXiv:2401.13649}, 2024.

\bibitem[Li et~al.(2020{\natexlab{a}})Li, Wang, Tang, Shi, Wu, Zhuang, and Wang]{li2020unsupervised}
Li, J., Wang, X., Tang, S., Shi, H., Wu, F., Zhuang, Y., and Wang, W.~Y.
\newblock Unsupervised reinforcement learning of transferable meta-skills for embodied navigation.
\newblock In \emph{Proceedings of the IEEE/CVF Conference on Computer Vision and Pattern Recognition}, pp.\  12123--12132, 2020{\natexlab{a}}.

\bibitem[Li et~al.(2022)Li, He, Wei, Qian, Zhu, Xie, Zhuang, Tian, and Tang]{li2022fine}
Li, J., He, X., Wei, L., Qian, L., Zhu, L., Xie, L., Zhuang, Y., Tian, Q., and Tang, S.
\newblock Fine-grained semantically aligned vision-language pre-training.
\newblock \emph{Advances in neural information processing systems}, 35:\penalty0 7290--7303, 2022.

\bibitem[Li et~al.(2023{\natexlab{a}})Li, Pan, Ge, Gao, Ji, Zhang, Chua, Tang, Zhang, and Zhuang]{li2023fine}
Li, J., Pan, K., Ge, Z., Gao, M., Ji, W., Zhang, W., Chua, T.-S., Tang, S., Zhang, H., and Zhuang, Y.
\newblock Fine-tuning multimodal llms to follow zero-shot demonstrative instructions.
\newblock In \emph{The Twelfth International Conference on Learning Representations}, 2023{\natexlab{a}}.

\bibitem[Li et~al.(2023{\natexlab{b}})Li, Tang, Zhu, Zhang, Yang, Chua, Wu, and Zhuang]{10121664}
Li, J., Tang, S., Zhu, L., Zhang, W., Yang, Y., Chua, T.-S., Wu, F., and Zhuang, Y.
\newblock Variational cross-graph reasoning and adaptive structured semantics learning for compositional temporal grounding.
\newblock \emph{IEEE Transactions on Pattern Analysis and Machine Intelligence}, 45\penalty0 (10):\penalty0 12601--12617, 2023{\natexlab{b}}.
\newblock \doi{10.1109/TPAMI.2023.3274139}.

\bibitem[Li et~al.(2020{\natexlab{b}})Li, He, Zhou, Zhang, and Baldridge]{ams}
Li, Y., He, J., Zhou, X., Zhang, Y., and Baldridge, J.
\newblock Mapping natural language instructions to mobile ui action sequences.
\newblock \emph{arXiv preprint arXiv:2005.03776}, 2020{\natexlab{b}}.

\bibitem[Lin et~al.(2024)Lin, Li, Gao, Yang, Wu, Bai, Lei, Wang, and Shou]{showui}
Lin, K.~Q., Li, L., Gao, D., Yang, Z., Wu, S., Bai, Z., Lei, W., Wang, L., and Shou, M.~Z.
\newblock Showui: One vision-language-action model for gui visual agent.
\newblock \emph{arXiv preprint arXiv:2411.17465}, 2024.

\bibitem[Lu et~al.(2024)Lu, Shao, Liu, Meng, Li, Chen, Huang, Zhang, Qiao, and Luo]{gui-odyssey}
Lu, Q., Shao, W., Liu, Z., Meng, F., Li, B., Chen, B., Huang, S., Zhang, K., Qiao, Y., and Luo, P.
\newblock Gui odyssey: A comprehensive dataset for cross-app gui navigation on mobile devices.
\newblock \emph{arXiv preprint arXiv:2406.08451}, 2024.

\bibitem[Miao et~al.(2025)Miao, Wu, Gao, Yu, Bu, Zhang, Li, Tang, Chua, and Li]{miao2025boostingvirtualagentlearning}
Miao, B., Wu, Y., Gao, M., Yu, Q., Bu, W., Zhang, W., Li, Y., Tang, S., Chua, T.-S., and Li, J.
\newblock Boosting virtual agent learning and reasoning: A step-wise, multi-dimensional, and generalist reward model with benchmark, 2025.
\newblock URL \url{https://arxiv.org/abs/2503.18665}.

\bibitem[Pan et~al.(2023)Pan, Li, Song, Lin, Liu, and Tang]{pan2023self}
Pan, K., Li, J., Song, H., Lin, J., Liu, X., and Tang, S.
\newblock Self-supervised meta-prompt learning with meta-gradient regularization for few-shot generalization.
\newblock \emph{arXiv preprint arXiv:2303.12314}, 2023.

\bibitem[Pan et~al.(2024{\natexlab{a}})Pan, Fan, Li, Yu, Fei, Tang, Hong, Zhang, and Sun]{pan2024towards}
Pan, K., Fan, Z., Li, J., Yu, Q., Fei, H., Tang, S., Hong, R., Zhang, H., and Sun, Q.
\newblock Towards unified multimodal editing with enhanced knowledge collaboration.
\newblock \emph{Advances in Neural Information Processing Systems}, 37:\penalty0 110290--110314, 2024{\natexlab{a}}.

\bibitem[Pan et~al.(2024{\natexlab{b}})Pan, Li, Wang, Fei, Song, Ji, Lin, Liu, Chua, and Tang]{pan2024i3}
Pan, K., Li, J., Wang, W., Fei, H., Song, H., Ji, W., Lin, J., Liu, X., Chua, T.-S., and Tang, S.
\newblock I3: I ntent-i ntrospective retrieval conditioned on i nstructions.
\newblock In \emph{Proceedings of the 47th International ACM SIGIR Conference on Research and Development in Information Retrieval}, pp.\  1839--1849, 2024{\natexlab{b}}.

\bibitem[Pan et~al.(2024{\natexlab{c}})Pan, Tang, Li, Fan, Chow, Yan, Chua, Zhuang, and Zhang]{pan2024auto}
Pan, K., Tang, S., Li, J., Fan, Z., Chow, W., Yan, S., Chua, T.-S., Zhuang, Y., and Zhang, H.
\newblock Auto-encoding morph-tokens for multimodal llm.
\newblock \emph{arXiv preprint arXiv:2405.01926}, 2024{\natexlab{c}}.

\bibitem[Pan et~al.(2025{\natexlab{a}})Pan, Lin, Yue, Ao, Jia, Zhao, Li, Tang, and Zhang]{pan2025generative}
Pan, K., Lin, W., Yue, Z., Ao, T., Jia, L., Zhao, W., Li, J., Tang, S., and Zhang, H.
\newblock Generative multimodal pretraining with discrete diffusion timestep tokens.
\newblock \emph{arXiv preprint arXiv:2504.14666}, 2025{\natexlab{a}}.

\bibitem[Pan et~al.(2025{\natexlab{b}})Pan, Wu, Bu, Shen, Li, Wang, Li, Tang, Xiao, Wu, et~al.]{pan2025unlocking}
Pan, K., Wu, Y., Bu, W., Shen, K., Li, J., Wang, Y., Li, Y., Tang, S., Xiao, J., Wu, F., et~al.
\newblock Unlocking aha moments via reinforcement learning: Advancing collaborative visual comprehension and generation.
\newblock \emph{arXiv preprint arXiv:2506.01480}, 2025{\natexlab{b}}.

\bibitem[Rawles et~al.(2024)Rawles, Li, Rodriguez, Riva, and Lillicrap]{aitw}
Rawles, C., Li, A., Rodriguez, D., Riva, O., and Lillicrap, T.
\newblock Androidinthewild: A large-scale dataset for android device control.
\newblock \emph{Advances in Neural Information Processing Systems}, 36, 2024.

\bibitem[Shen et~al.(2024)Shen, Jain, Xiao, Amlekar, Hadji, Podolny, and Talwalkar]{scribeagent}
Shen, J., Jain, A., Xiao, Z., Amlekar, I., Hadji, M., Podolny, A., and Talwalkar, A.
\newblock Scribeagent: Towards specialized web agents using production-scale workflow data.
\newblock \emph{arXiv preprint arXiv:2411.15004}, 2024.

\bibitem[Shen et~al.(2023)Shen, Song, Tan, Zhang, Ren, Yuan, Lu, Li, and Zhuang]{taskbench}
Shen, Y., Song, K., Tan, X., Zhang, W., Ren, K., Yuan, S., Lu, W., Li, D., and Zhuang, Y.
\newblock Taskbench: Benchmarking large language models for task automation.
\newblock \emph{arXiv preprint arXiv:2311.18760}, 2023.

\bibitem[Wang et~al.(2024{\natexlab{a}})Wang, Deng, Zha, Mao, Wang, Min, Chen, and Chen]{mobileagentbench}
Wang, L., Deng, Y., Zha, Y., Mao, G., Wang, Q., Min, T., Chen, W., and Chen, S.
\newblock Mobileagentbench: An efficient and user-friendly benchmark for mobile llm agents.
\newblock \emph{arXiv preprint arXiv:2406.08184}, 2024{\natexlab{a}}.

\bibitem[Wang et~al.(2024{\natexlab{b}})Wang, Bai, Tan, Wang, Fan, Bai, Chen, Liu, Wang, Ge, Fan, Dang, Du, Ren, Men, Liu, Zhou, Zhou, and Lin]{qwen2vl}
Wang, P., Bai, S., Tan, S., Wang, S., Fan, Z., Bai, J., Chen, K., Liu, X., Wang, J., Ge, W., Fan, Y., Dang, K., Du, M., Ren, X., Men, R., Liu, D., Zhou, C., Zhou, J., and Lin, J.
\newblock Qwen2-vl: Enhancing vision-language model's perception of the world at any resolution.
\newblock \emph{arXiv preprint arXiv:2409.12191}, 2024{\natexlab{b}}.

\bibitem[Wu et~al.(2024{\natexlab{a}})Wu, Fei, Qu, Ji, and Chua]{wu24next}
Wu, S., Fei, H., Qu, L., Ji, W., and Chua, T.-S.
\newblock {NE}x{T}-{GPT}: Any-to-any multimodal {LLM}.
\newblock In \emph{Proceedings of the International Conference on Machine Learning}, pp.\  53366--53397, 2024{\natexlab{a}}.

\bibitem[Wu et~al.(2024{\natexlab{b}})Wu, Wu, Xu, Wang, Sun, Jia, Cheng, Ding, Chen, Liang, et~al.]{os-atlas}
Wu, Z., Wu, Z., Xu, F., Wang, Y., Sun, Q., Jia, C., Cheng, K., Ding, Z., Chen, L., Liang, P.~P., et~al.
\newblock Os-atlas: A foundation action model for generalist gui agents.
\newblock \emph{arXiv preprint arXiv:2410.23218}, 2024{\natexlab{b}}.

\bibitem[Xie et~al.(2024)Xie, Zhang, Chen, Li, Zhao, Cao, Hua, Cheng, Shin, Lei, et~al.]{osworld}
Xie, T., Zhang, D., Chen, J., Li, X., Zhao, S., Cao, R., Hua, T.~J., Cheng, Z., Shin, D., Lei, F., et~al.
\newblock Osworld: Benchmarking multimodal agents for open-ended tasks in real computer environments.
\newblock \emph{arXiv preprint arXiv:2404.07972}, 2024.

\bibitem[Xu et~al.(2024{\natexlab{a}})Xu, Chen, Wu, Chen, Zhang, Yao, Xie, Chen, Liu, Qian, et~al.]{crab}
Xu, T., Chen, L., Wu, D.-J., Chen, Y., Zhang, Z., Yao, X., Xie, Z., Chen, Y., Liu, S., Qian, B., et~al.
\newblock Crab: Cross-environment agent benchmark for multimodal language model agents.
\newblock \emph{arXiv preprint arXiv:2407.01511}, 2024{\natexlab{a}}.

\bibitem[Xu et~al.(2024{\natexlab{b}})Xu, Wang, Wang, Lu, Xie, Saha, Sahoo, Yu, and Xiong]{aguvis}
Xu, Y., Wang, Z., Wang, J., Lu, D., Xie, T., Saha, A., Sahoo, D., Yu, T., and Xiong, C.
\newblock Aguvis: Unified pure vision agents for autonomous gui interaction.
\newblock \emph{arXiv preprint arXiv:2412.04454}, 2024{\natexlab{b}}.

\bibitem[Yao et~al.(2022)Yao, Chen, Yang, and Narasimhan]{webshop}
Yao, S., Chen, H., Yang, J., and Narasimhan, K.
\newblock Webshop: Towards scalable real-world web interaction with grounded language agents.
\newblock \emph{Advances in Neural Information Processing Systems}, 35:\penalty0 20744--20757, 2022.

\bibitem[Zhou et~al.(2023)Zhou, Xu, Zhu, Zhou, Lo, Sridhar, Cheng, Ou, Bisk, Fried, et~al.]{webarena}
Zhou, S., Xu, F.~F., Zhu, H., Zhou, X., Lo, R., Sridhar, A., Cheng, X., Ou, T., Bisk, Y., Fried, D., et~al.
\newblock Webarena: A realistic web environment for building autonomous agents.
\newblock \emph{arXiv preprint arXiv:2307.13854}, 2023.

\end{thebibliography}
\bibliographystyle{icml2025}

\addtocontents{toc}{\protect\setcounter{tocdepth}{2}}
\newpage
\onecolumn
\appendix
\renewcommand{\contentsname}{\textbf{Table of Contents in Appendix}} 
\renewcommand{\cftsecfont}{\large\bfseries} 
\renewcommand{\cftsecpagefont}{\bfseries}
\renewcommand{\cftsubsecfont}{\large} 
\renewcommand{\cftsecleader}{\cftdotfill{\cftdotsep}} 
\setlength{\cftbeforesecskip}{20pt}
\setlength{\cftbeforesubsecskip}{8pt}
\renewcommand{\thesection}{\Alph{section}}
\renewcommand{\thesubsection}{\thesection.\arabic{subsection}}
 
\renewcommand{\contentsname}{Table of Contents in Appendix}
\hypersetup{linkcolor=black}
\tableofcontents
\clearpage
\hypersetup{linkcolor=blue}

\twocolumn

\section{Environment Setup}
OmniBench conducts agent evaluations across three categories of virtual environments: Desktop, Mobile, and Web. Its automated data collection pipeline makes it easy to extend the benchmark to additional environments with minimal effort. In the following section, we use the desktop environment as a case study to illustrate the environment design in OmniBench in more detail.

\subsection{Environment Infrastructure}
Inspired by OSWorld~\cite{osworld}, we design an interactive Windows-based environment using a virtual machine to support GUI agent evaluation. The environment runs Windows 11 as the guest operating system and uses VMware Workstation 17 Pro (version 17.5.1) as the virtualization platform. This setup enables high compatibility with real-world desktop applications while maintaining full control over the execution environment. The virtual machine allows us to simulate user interactions such as mouse clicks, keyboard input, and file operations, which are essential for GUI agents. It also supports real-time observation and logging of system states, facilitating fine-grained analysis and reproducibility of agent behavior. All environments are initialized from a snapshot to ensure consistent starting conditions for each evaluation episode. 

\subsection{Observation Space}
\label{observation_space}

In OmniBench, the observation space is designed to ensure comprehensive evaluation of GUI agents by capturing both visual and structural aspects of desktop environments. It comprises two complementary modalities: screen captures and accessibility trees. This dual-modality approach reflects the varying grounding capabilities of different agent architectures. For example, agents that have been specifically trained on GUI environments often possess strong grounding abilities and can rely on screen captures alone. In contrast, MLLMs typically lack specialized pretraining for GUI understanding, and therefore benefit significantly from the semantic and structural information provided by the accessibility tree. By supporting both modalities, OmniBench enables fair and informative evaluation across a wide range of agents, ensuring robust evaluation under diverse UI layouts and application contexts.

\subsection{Action Space}
\label{action_space}
In OmniBench, the action space consists of three core types of user interactions that an agent can perform. These actions, summarized in Table~\ref{tab:action_space}, enable the agent to effectively interact with graphical user interfaces across a wide range of applications.

\begin{table}[h]
\centering
\caption{Summary of action types in the desktop environment of OmniBench.}
\label{tab:action_space}
\resizebox{\linewidth}{!}{
\begin{tabular}{p{4cm}|p{10cm}}
\toprule
\textbf{Action} & \textbf{Description} \\
\midrule
\texttt{click\_input} & Simulates mouse clicks on UI control elements. Supports configurable mouse buttons (left, right, middle, x) and can perform both single and double clicks. Commonly used for selecting items, activating controls, or opening folders. \\
\midrule
\texttt{wheel\_mouse\_input} & Scrolls vertically using the mouse wheel. Useful when target controls are not immediately visible. The scroll direction and distance are adjustable, allowing the agent to navigate long content or lists. \\
\midrule
\texttt{keyboard\_input} & Simulates keyboard input for typing text, pressing keys, or invoking shortcuts (e.g., \texttt{Ctrl+C}, \texttt{Enter}). Enables fine-grained control over application behavior and supports both functional and textual input. \\
\bottomrule
\end{tabular}
}
\end{table}

\subsection{Categorization of Applications}
\label{app_cate}
The 49 applications in the environment are categorized into 12 distinct groups based on their functionality: Social Communication, Multimedia Playback, Multimedia Editing, Office, Utility Tools, Programming, System Management, Web Browsing, Screen Capture, Task Management, Note Management, and Lifestyle. The specific applications belonging to each category are listed in Table~\ref{tab:app_categories}.

\begin{table}[ht]
    \centering
    \caption{Categorization of applications in the environment.}
    \resizebox{\linewidth}{!}{
    \begin{tabular}{p{4cm}|p{8cm}}
        \toprule
        \textbf{Category} & \textbf{Applications} \\
        \midrule
        Social Communication~(4) & Zoom Workplace, Skype, People, Mail \\
        Multimedia Playback~(4) & Media Player, Spotify, Photos, TuneIn \\
        Multimedia Editing~(6) & Adobe Photoshop Express, Microsoft Clipchamp, paint.net, Openshot, Handbrake, Paint \\
        Office~(3) & Word, PowerPoint, Excel \\
        Utility Tools~(10) & Calculator, 7-Zip, PDF24, Power Automate, Wikipedia, BreeZip, Maps, Calendar, Zotero, DeepL \\
        Programming~(3) & Visual Studio Code, Cursor, Windows PowerShell ISE \\
        System Management~(4) & File Explorer, Settings, Control Panel, Microsoft Store \\
        Web Browsing~(2) & Google Chrome, Microsoft Edge \\
        Screen Capture~(4) & Record Screen, Snipping Tool, OBS Studio, ShareX \\
        Task Management~(3) & Microsoft To Do, Todoist, Notion \\
        Note management~(4) & Evernote, OneNote, Sticky Notes, Sticky Notes~(New)\\
        Lifestyle~(2) & Recipe Keeper, paisa \\
        \bottomrule
    \end{tabular}
    }
    \label{tab:app_categories}
\end{table}

\subsection{Definition of Scenarios}

\begin{figure}[h!]
\includegraphics[width=\linewidth]{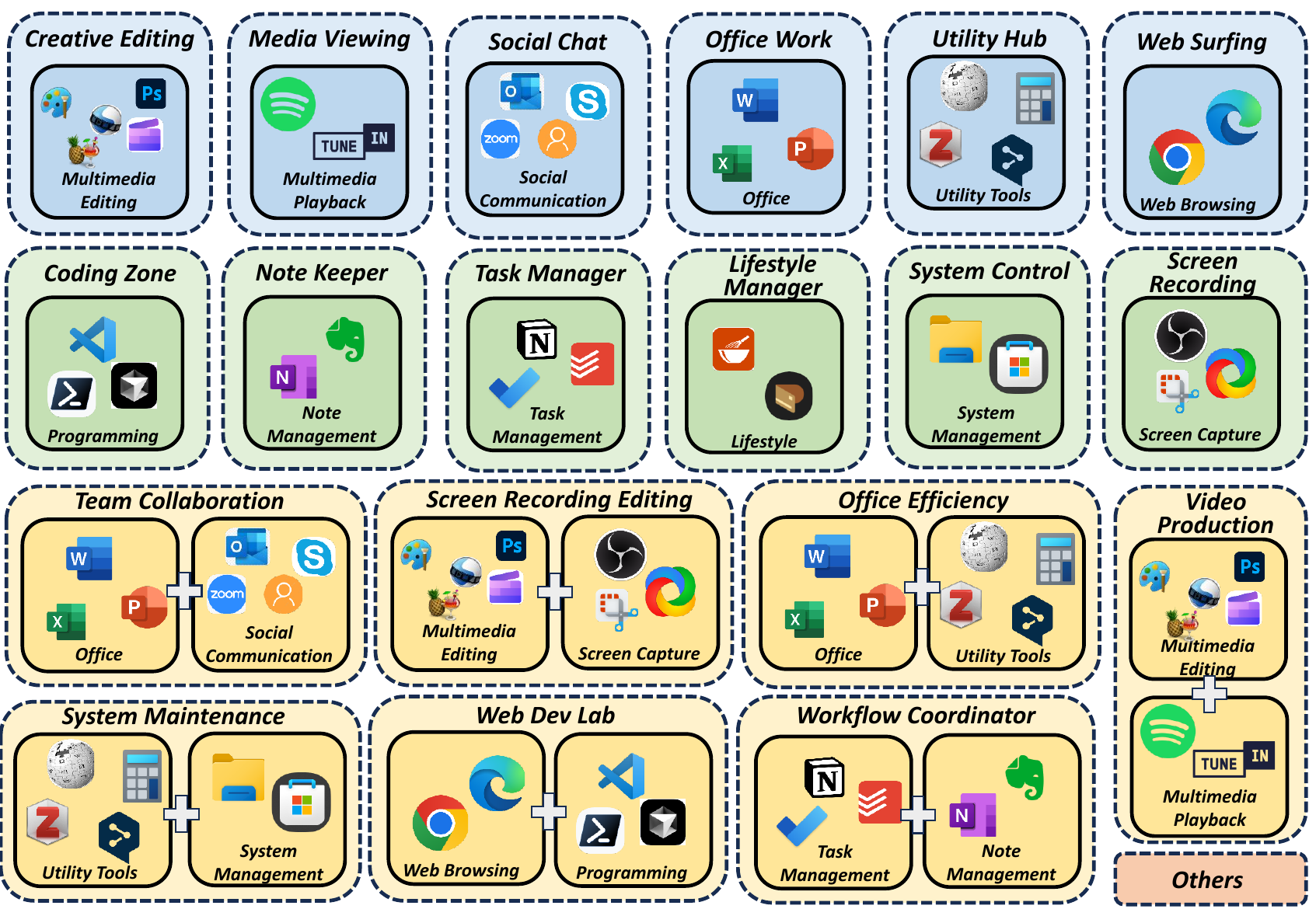}
\centering\caption{Overview of the 20 defined task scenarios. The scenarios are derived from combinations of application categories, where each box represents a task scenario with associated application icons.}
\label{scenario}
\end{figure}

As shown in Table~\ref{tab:app_categories}, we categorize the 49 applications into 12 groups. Based on these application categories, we define 12 corresponding task scenarios. For example, applications in the ``Multimedia Playback'' category correspond to the ``Media Viewing'' scenario. We then combine these 12 basic task scenarios to form 7 more fine-grained task scenarios. For instance, the ``Screen Recording'' scenario can be combined with the ``Creative Editing'' scenario to form a more detailed scenario called ``Screen Recording Editing.'' Through such combinations, we obtain a total of 19 task scenarios. The remaining tasks are grouped into an additional scenario, resulting in a final set of 20 task scenarios, as illustrated in Figure~\ref{scenario}.

\section{Data Collection}
OmniBench introduces a bottom-up pipeline for automatic task synthesis, as illustrated in Figure~\ref{pipeline}. The pipeline consists of two main stages: Subtask Synthesis and Task Synthesis. The Subtask Synthesis stage includes two steps: Subtask Discovery and Iterative Synthesis, while the Task Synthesis stage consists of Task Composition and Task Validation. In the following sections, we provide a detailed explanation of each component in this data synthesis pipeline.

\subsection{Subtask Discovery}
\label{A3.2.1}
To facilitate comprehensive subtask generation, we construct a dynamic environment that incorporates 49 heterogeneous applications, drawing inspiration from OSWorld~\cite{osworld}. This environment enables MLLMs to interact with diverse functionalities, systematically analyze operational constraints, and propose well-structured subtasks tailored to each application's context. To support this process, we provide detailed API documentation, user manuals, and curated example subtasks, ensuring that MLLMs can infer practical usage scenarios. In addition, our approach emphasizes dependency modeling through a structured resource framework, where each subtask explicitly defines its required inputs and expected outputs. This predefined resource list serves as a guiding constraint, allowing MLLMs to reason about inter-subtask dependencies, avoid conflicts, and ensure smooth task execution. By leveraging this controlled exploration, MLLMs generate coherent and executable subtask sequences that align with real-world application workflows.

\subsection{Iterative Synthesis}
\label{cross}

\textbf{Trajectory Synthesis.} 
With the rapid development of MLLMs~\cite{li2022fine, pan2025generative, pan2024towards, pan2023self}, they are becoming increasingly capable~\cite{li2020unsupervised, pan2024auto}. We deploy state-of-the-art MLLMs to execute the synthesized subtasks and record their execution traces as synthesized trajectories. Since subtasks are typically composed of a sequence of relatively simple instructions, advanced MLLMs achieve high success rates on these subtasks, making them suitable for generating an initial set of trajectories.

\begin{figure}[h!]
\includegraphics[width=\linewidth]{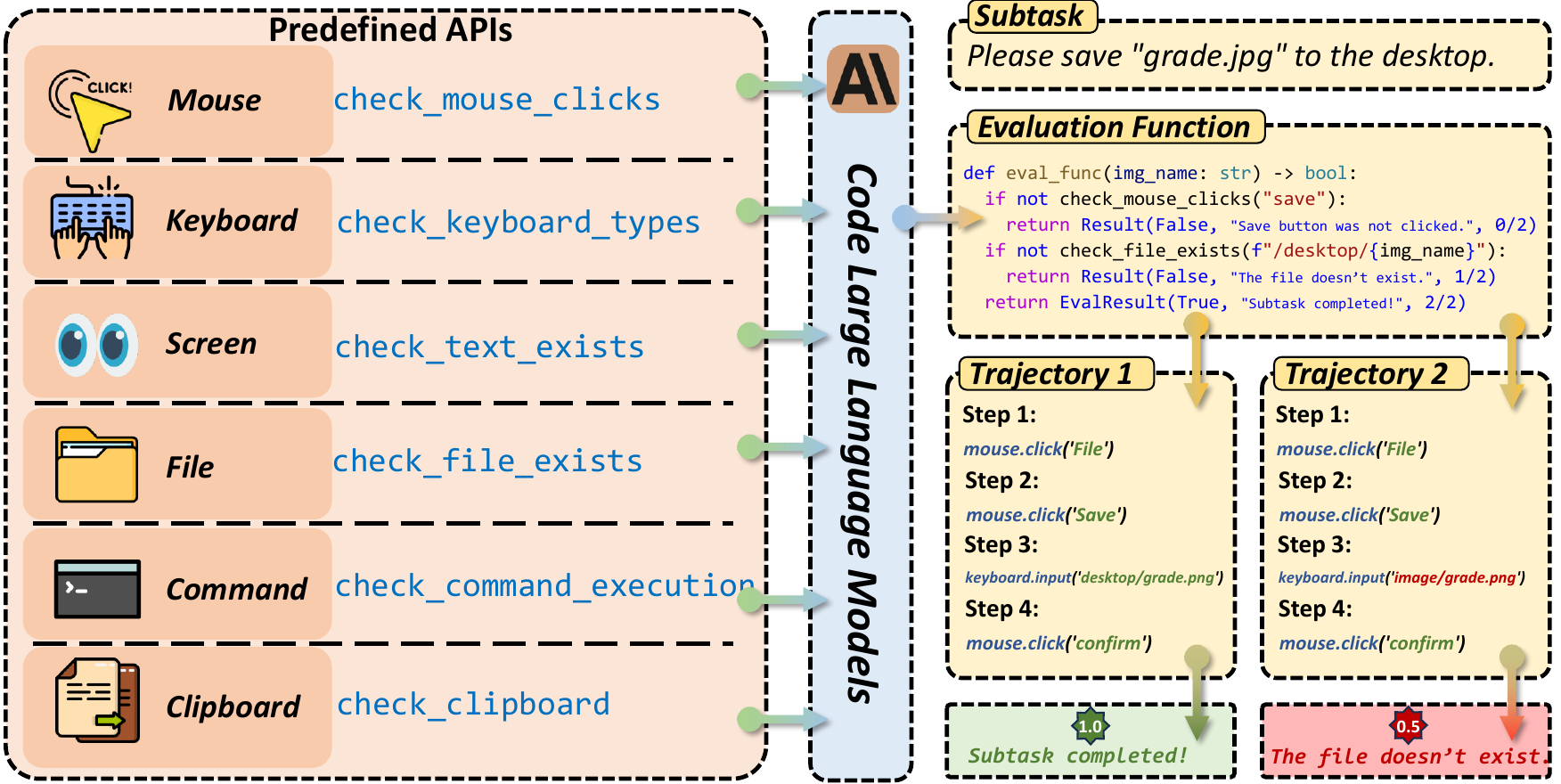}
\centering\caption{An overview of how predefined system-level APIs are composed into evaluation functions using Code LLMs.}
\label{API2func}
\end{figure}

\textbf{Evaluation Synthesis.}
Inspired by prior work~\cite{minghegao_mm1, gao2025benchmarkingmultimodalcotreward, fei2024enhancing, fei2024video, 10121664, pan2025unlocking}, we first manually designed 11 system-level APIs, such as retrieving the text under the mouse cursor, accessing the clipboard content, and extracting visible text from the screen. We then provided the function signatures of these APIs to Code LLMs, which generated evaluation functions used to determine the completion status of sub-tasks. Since these evaluation functions are composed by invoking the designed APIs, we are able to assign more fine-grained evaluation scores based on the number of successfully executed API calls. Figure~\ref{API2func} illustrates the process of composing sub-task evaluation functions using these APIs.

\textbf{Cross-Verification.} We design a cross-verification algorithm to optimize synthesized subtasks. Specifically, the algorithm performs \( N \) iterations, where an MLLM and a Code LLM specifically synthesize trajectories and evaluation functions based on the feedback from the previous iteration. The evaluation functions provide detailed failure feedback (\textit{e.g.}, failing to click the ``save" button), while the trajectories incorporate environmental state feedback, enabling an iterative refinement process for both trajectories and evaluation functions. To ensure the evaluation functions maintain sufficient discriminatory power, any function that passes cross-verification is further tested by evaluating three additional trajectories from other tasks. 

\subsection{Task Composition}
\label{A3.2.3}
\begin{figure}[h!]
\includegraphics[width=\linewidth]{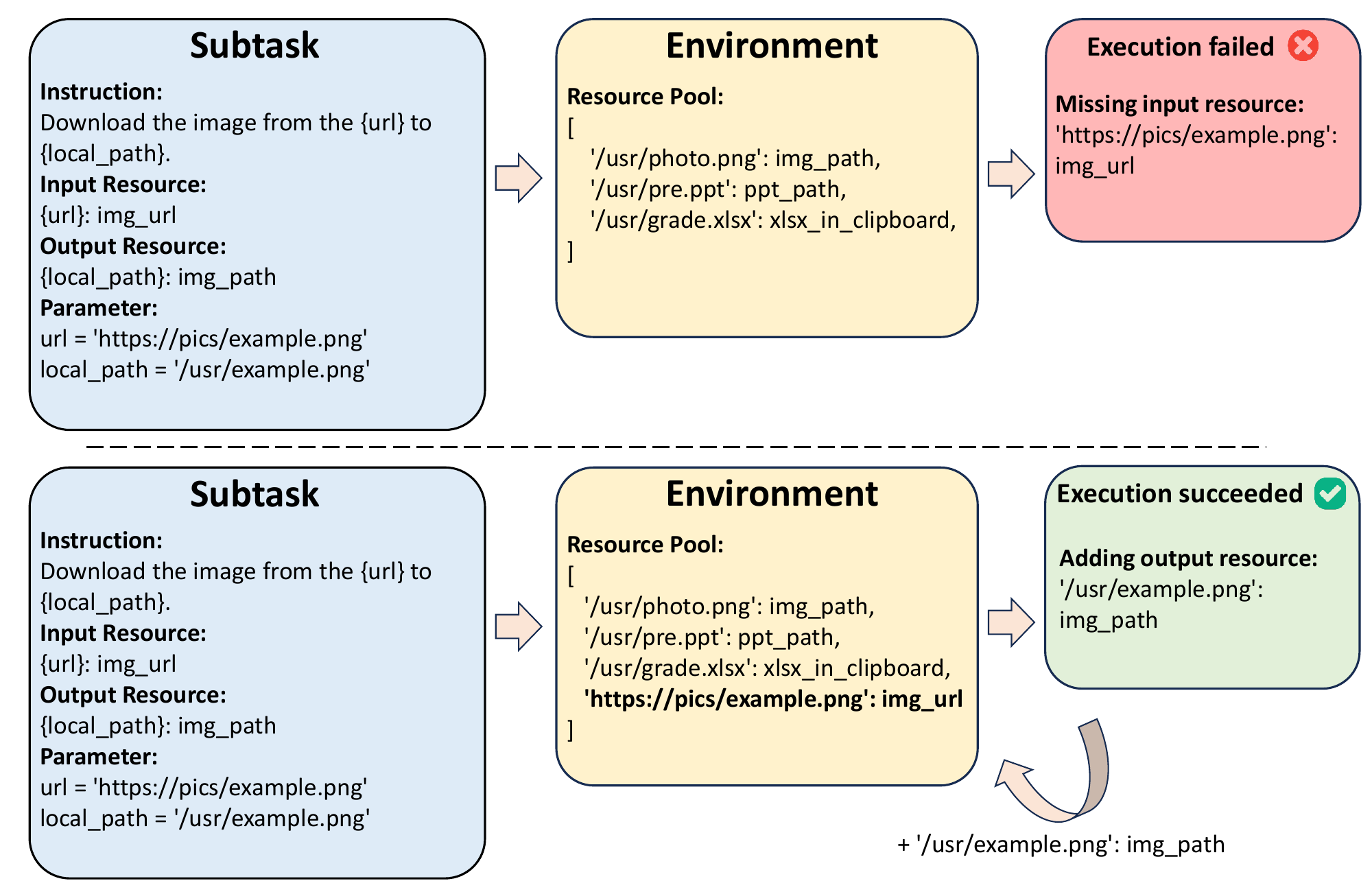}
\centering\caption{An overview of the constraint relationships among environmental resources.}
\label{resource}
\end{figure}

\noindent \textbf{Environmental resources for task composition.}
To precisely capture the execution dependencies between different subtasks, we propose a set of concepts for environmental resources, as shown in Figure~\ref{resource}. Specifically, each environmental resource has a resource category and an actual parameter. For example, we can use \texttt{img\_path} to represent a category of images that exist locally, and the actual parameter \texttt{/usr/example.png} then instantiates this environmental resource. Each subtask has an input resource list and an output resource list, which represent the prerequisite resources required to execute the subtask and the new resources generated after execution, respectively. Through this method of resource representation and the logic of resource transformation, we can clearly define the dependencies between subtasks. A dependency relationship exists between two subtasks only if one subtask can provide the resources in the input resource list of the other subtask.

\noindent \textbf{Task Composition with Consistent Intent.} To construct meaningful task graphs, we first curate a subtask pool by filtering out low-quality samples through a cross-verification process. A naïve bottom-up approach that directly connects subtasks based on shared input-output resources may generate incoherent tasks that lack a clear, unified goal. For example, an arbitrary combination might result in a task that simultaneously plays media from different applications without a meaningful relationship. To mitigate this issue, we explicitly extract overarching task intents from the subtask pool, such as `create a personal introduction PowerPoint for Emily," as illustrated in Figure~\ref{mainfigure}. Each intent acts as an anchor, grouping subtasks that contribute to the same high-level objective. The task graph is then constructed by linking these subtasks based on resource dependencies. Since the bottom-up composition follows predefined rules, we maintain strict control over the synthesis process, ensuring that the resulting task graphs exhibit logical coherence and adjustable complexity. This structured approach prevents the generation of ill-formed tasks while supporting diverse and meaningful compositions.

\subsection{Task Validation}
\label{A3.2.4}
To generate concise yet accurate task instructions for graph-structured tasks, we utilize GPT-4o to synthesize a task description by integrating subtask instructions and the structural information of the task graph. However, direct summarization may introduce inconsistencies, such as misrepresenting the graph’s dependency structure or oversimplifying it into a linear sequence, thereby distorting the original task complexity. To address this issue, we designed a consistency validation mechanism to evaluate the fidelity of the summarized instruction. In this process, GPT-4o is prompted to infer subtask dependencies purely from the generated task instruction, without access to the original task graph. The inferred dependency structure is then compared against the ground-truth graph. If discrepancies arise—such as missing parallel execution paths or incorrect sequencing—the instruction is flagged for revision and must be re-summarized to better preserve the intended graph structure. This validation ensures that the final task instruction accurately reflects the hierarchical and branching nature of the original task graph, improving the clarity and correctness of the synthesized tasks.

\section{Details of OmniBench}
In this section, we provide a detailed description of the data formats used in OmniBench. The data in OmniBench can be broadly categorized into five types: Subtask Metadata, Subtask Trajectory, Subtask Evaluation, Task Metadata, and Task Trajectory. For each category, we present the corresponding data schema along with representative examples. Finally, we provide a visualization example of a task graph.

\subsection{Subtask Metadata}
Each subtask includes the following seven attributes:
\begin{itemize}
  \item \textbf{id}: A UUID that uniquely identifies the subtask.
  \item \textbf{instruction\_template}: A template of the subtask instruction containing parameter placeholders.
  \item \textbf{application}: The name of the application to which the subtask belongs.
  \item \textbf{available\_parameters}: A list of all configurable parameter sets used to instantiate diverse instructions. Each element in the list represents one valid combination of parameters.
  \item \textbf{OS}: The operating system associated with the subtask.
  \item \textbf{input\_resources}: A list of prerequisite resources required before executing the subtask.
  \item \textbf{output\_resources}: A list of resources produced after the subtask is completed.
\end{itemize}

\begin{tcolorbox}[
    colback=gray!5,
    colframe=blue!60!black,
    title=Example of Subtask Metadata,
    listing only,
    breakable]
\begin{minted}[fontsize=\footnotesize, breaklines=true, breakanywhere=true, breaksymbolleft={}, breaksymbolright={}]{json}
{
    "id": "25e2a51e-c019-1a9a-0747-d6fe0e9d457d",
    "instruction_template": "Open '{xlsx_path}', select the all data, and copy it.",
    "application": "Excel",
    "available_parameters": [
        {
            "xlsx_path": "C:\\Users\\user\\Desktop\\office\\The Evolution of Urbanization Rate.xlsx"
        }
    ],
    "OS": "Windows",
    "input_resources": [
        "xlsx_path"
    ],
    "output_resources": [
        "table_in_clipboard",
        "xlsx_in_processing"
    ]
}
\end{minted}
\end{tcolorbox}

\subsection{Subtask Trajectory}
Each subtask trajectory includes the following five attributes:
\begin{itemize}
  \item \textbf{trajectory\_id}: A unique identifier for the trajectory.
  \item \textbf{instruction}: The instantiated instruction, generated by applying a specific parameter set from the subtask's \texttt{available\_parameters}.
  \item \textbf{observations}: A list of screenshots representing each step in the trajectory.
  \item \textbf{actions}: A list of actions taken at each step of the trajectory.
  \item \textbf{subtask\_id}: The identifier of the subtask to which this trajectory corresponds.
\end{itemize}

\begin{tcolorbox}[colback=gray!5, colframe=blue!60!black, title=Example of Subtask Trajectory, listing only, breakable]
\begin{minted}[fontsize=\footnotesize, breaklines=true, breakanywhere=true, breaksymbolleft={}, breaksymbolright={}]{json}
{
    "trajectory_id": "XXX",
    "instruction": "Using the file explorer, navigate to C:\\Users\\user\\Desktop\\images\\ and new a Text Document named introduction.txt",
    "observations": [
        "obs1.png", "obs2.png", "..."
    ],
    "actions": [
        {
            "function": "click_input",
            "args": {
                "button": "left",
                "double": false
            },
            "rect": [
                124,
                1020,
                179,
                1080
            ],
            "description": "There are many application icons on the taskbar, and I need to select the File Explorer to complete the task.",
            "thought": "To fulfill 'Using the file explorer, navigate to C:\\Users\\user\\Desktop\\images\\ and new a Text Document named introduction.txt', I need to first click the 'File Explorer' button to open the corresponding application.",
            "control_text": "File Explorer"
        }, 
        {
          "function": "...",
          "args": { "...": "..." }
        }
    ],
    "subtask_id": "XXX"
}

\end{minted}
\end{tcolorbox}

\subsection{Subtask Evaluation} Each subtask evaluation is implemented as a Python function that checks for a sequence of expected interaction outcomes based on predefined system-level APIs. The function returns an \texttt{EvalResult} object containing three fields:

\begin{itemize}
  \item \textbf{success}: A boolean value indicating whether the subtask was completed successfully.
  \item \textbf{message}: A string providing a human-readable explanation of the evaluation result.
  \item \textbf{progress}: A float or fraction representing the proportion of evaluation conditions that were satisfied, enabling partial credit.
\end{itemize}

Each evaluation function typically performs multiple checks, such as verifying whether specific UI elements were clicked, expected text appeared on the screen, or input was typed correctly. These checks are implemented using task-agnostic API functions (e.g., \texttt{check\_mouse\_clicks}, \texttt{check\_text\_exists\_via\_control}, \texttt{check\_keyboard\_types}). By combining these low-level signals, the evaluation function provides a fine-grained and automated judgment of subtask execution.

\begin{tcolorbox}[
  colback=gray!5,
  colframe=blue!60!black,
  title=Example of Subtask Evaluation,
  listing only,
  breakable
]
\begin{minted}[
  fontsize=\footnotesize,
  breaklines=true,
  breakanywhere=true,
  breaksymbolleft={},
  breaksymbolright={}
]{python}
from collections import namedtuple

EvalResult = namedtuple('EvalResult', ['success', 'message', 'progress'])

def evaluate_agent_task_completion(csv_path: str) -> EvalResult:
    if not check_mouse_clicks(text='More actions'):
        return EvalResult(False, "Subtask execution failed because agent did not click the 'More actions' button.", 0/4)

    if not check_text_exists_via_control(text='Import tasks from a spreadsheet using a CSV file.'):
        return EvalResult(False, "Subtask execution failed because the import tasks option was not accessed.", 1/4)

    if not check_keyboard_types(text=csv_path):
        return EvalResult(False, f"Subtask execution failed because the CSV file path '{csv_path}' was not typed.", 2/4)

    if not check_mouse_clicks(text='Open'):
        return EvalResult(False, "Subtask execution failed because the 'Open' button was not clicked to import the file.", 3/4)

    return EvalResult(True, "Subtask completed successfully", 4/4)
\end{minted}
\end{tcolorbox}

\subsection{Task Metadata}

Each task includes the following four attributes:
\begin{itemize}
  \item \textbf{task\_instruction}: The natural language instruction that describes the overall task.
  \item \textbf{dag}: A directed acyclic graph (DAG) representing the structural dependencies among subtasks within the task.
  \item \textbf{task\_intent}: The high-level goal or intent that the task is designed to achieve.
  \item \textbf{successful\_topo}: All valid topological orders of the DAG that lead to successful task completion.
\end{itemize}

\begin{tcolorbox}[colback=gray!5, colframe=blue!60!black, title=Example of Task Metadata, listing only, breakable]
\begin{minted}[fontsize=\footnotesize, breaklines=true, breakanywhere=true, breaksymbolleft={}, breaksymbolright={}]{json}
{
    "task_instruction": "In Excel, open 'C:\\Users\\user\\Desktop\\office\\The Evolution of Urbanization Rate.xlsx', select the 'A' column, and center the content. Then, export the document as a PDF named 'C:\\Users\\user\\Desktop\\pdf\\The Evolution of Urbanization Rate.pdf'.",
    "dag": {
        "nodes": [
            "a7310aa0-b194-77e3-5c36-996391a1bc7d",
            "df3fc68b-fa76-4e19-7da6-aef17792523b"
        ],
        "edges": {
            "a7310aa0-b194-77e3-5c36-996391a1bc7d": [
                "df3fc68b-fa76-4e19-7da6-aef17792523b"
            ],
            "df3fc68b-fa76-4e19-7da6-aef17792523b": []
        }
    },
    "task_intent": "Center Excel data and export to PDF",
    "successful_topo": [
        [
            "a7310aa0-b194-77e3-5c36-996391a1bc7d",
            "df3fc68b-fa76-4e19-7da6-aef17792523b"
        ]
    ]
}
\end{minted}
\end{tcolorbox}

\subsection{Task Trajectory}
Each task trajectory includes the following seven attributes:
\begin{itemize}
  \item \textbf{trajectory\_id}: The unique identifier of the trajectory. The suffix (0) indicates that it corresponds to the first topological order in the \texttt{successful\_topo} list.
  \item \textbf{task\_id}: The identifier of the associated task.
  \item \textbf{topological\_order}: The specific topological order followed in this trajectory.
  \item \textbf{instruction}: The instruction describing the overall task.
  \item \textbf{intent}: The high-level intent or goal of the task.
  \item \textbf{observations}: The sequence of visual observations recorded during task execution.
  \item \textbf{actions}: The sequence of actions taken during task execution.
\end{itemize}

\begin{tcolorbox}[colback=gray!5, colframe=blue!60!black, title=Example of Task Trajectory, listing only, breakable]
\begin{minted}[fontsize=\footnotesize, breaklines=true, breakanywhere=true, breaksymbolleft={}, breaksymbolright={}]{json}
{
  "trajectory_id": "12(0)",
  "task_id": "12",
  "topological_order": [
    "a7310aa0-b194-77e3-5c36-996391a1bc7d",
    "df3fc68b-fa76-4e19-7da6-aef17792523b"
  ],
  "instruction": "In Excel, open the file, center the A column, and export as PDF.",
  "intent": "Center Excel data and export to PDF",
  "observations": [
    "obs1.png", "obs2.png", "..."
  ],
  "actions": [
    {
      "function": "click_input",
      "args": {
        "button": "left",
        "double": true
      },
      "rect": [1520, 371, 1614, 458],
      "description": "Double-click the 'Excel' icon on the desktop.",
      "thought": "To begin the task, I need to open Excel.",
      "control_text": "Excel"
    },
    {
      "function": "...",
      "args": { "...": "..." }
    }
  ]
}

\end{minted}
\end{tcolorbox}

\subsection{Example of Task Graph}
\begin{figure}[h!]
\includegraphics[width=\linewidth]{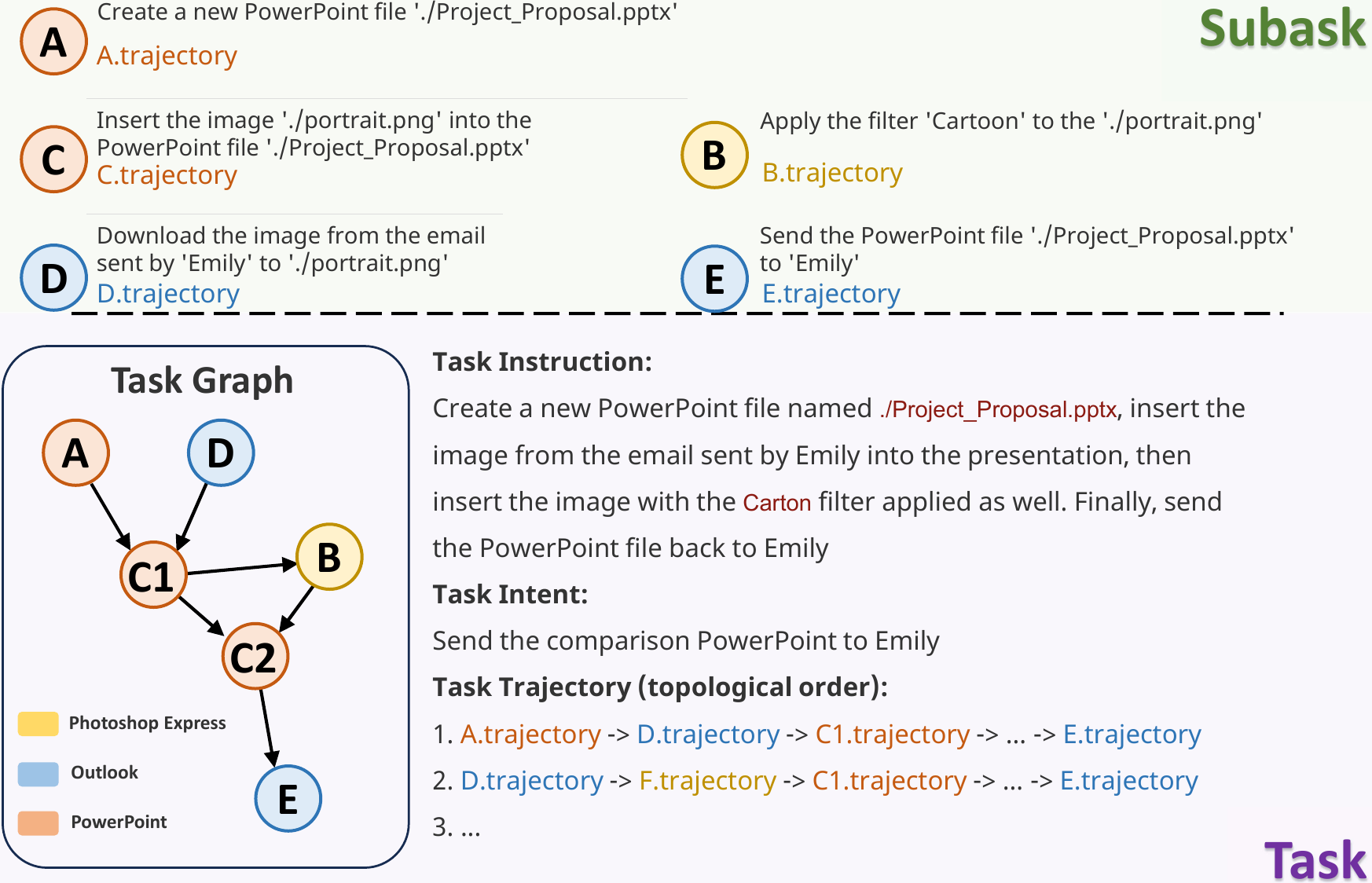}
\centering\caption{A task example in OmniBench.}
\label{case}
\end{figure}

To more clearly illustrate the relationships between subtasks and tasks, we provide an additional visualization example of a task graph, as shown in Figure~\ref{case}. Each node in the graph signifies a distinct subtask, categorized by the application used: PowerPoint (orange), Outlook (blue), and Photoshop Express (yellow). The directed edges denote the sequence of execution, where each subtask must be completed before the subsequent ones can begin.

\section{Details of OmniEval}
In this section, we introduce the details of OmniEval, including the design of ten essential capabilities, the experimental setup for evaluation, and implementation details during the evaluation.

\subsection{Capability Design}
\label{compl_and_cap}
As shown in Table~\ref{tab:cap}, we construct 10 specialized sets of test tasks based on the five-dimensional complexity, each set designed to evaluate one of the 10 capabilities required for agents to fulfill user requests. The number of solid stars represents the level of complexity, while zero stars indicate that the complexity in that dimension can be arbitrary.

\begin{table}[h!]
\centering
\resizebox{\linewidth}{!}{
\begin{tabular}{l|ccccc}
\hline
\textbf{} & Dependency & Instruction & Hierarchy & Branch & Knowledge \\
\hline
Parallel Planning & $\solidstar\solidstar\solidstar$ & $\hollowstar\hollowstar\hollowstar$ & $\hollowstar\hollowstar\hollowstar$ & $\solidstar\solidstar\solidstar$ & $\hollowstar\hollowstar\hollowstar$ \\ 
Long-Range Planning & $\solidstar\solidstar\solidstar$ & $\hollowstar\hollowstar\hollowstar$ & $\solidstar\solidstar\solidstar$ & $\hollowstar\hollowstar\hollowstar$ & $\hollowstar\hollowstar\hollowstar$ \\ 
Long-Sequence Reasoning & $\hollowstar\hollowstar\hollowstar$ & $\solidstar\solidstar\solidstar$ & $\solidstar\solidstar\solidstar$ & $\hollowstar\hollowstar\hollowstar$ & $\hollowstar\hollowstar\hollowstar$ \\ 
Long Instruction Following & $\hollowstar\hollowstar\hollowstar$ & $\hollowstar\hollowstar\hollowstar$ & $\solidstar\hollowstar\hollowstar$ & $\solidstar\solidstar\solidstar$ & $\hollowstar\hollowstar\hollowstar$ \\ 
Sequential Decision-Making & $\hollowstar\hollowstar\hollowstar$ & $\hollowstar\hollowstar\hollowstar$ & $\solidstar\solidstar\solidstar$ & $\solidstar\solidstar\solidstar$ & $\hollowstar\hollowstar\hollowstar$ \\ 
Cross-Domain Decision-Making & $\hollowstar\hollowstar\hollowstar$ & $\hollowstar\hollowstar\hollowstar$ & $\hollowstar\hollowstar\hollowstar$ & $\solidstar\solidstar\solidstar$ & $\solidstar\solidstar\solidstar$ \\ 
Subtask Identification & $\solidstar\hollowstar\hollowstar$ & $\solidstar\solidstar\solidstar$ & $\hollowstar\hollowstar\hollowstar$ & $\hollowstar\hollowstar\hollowstar$ & $\hollowstar\hollowstar\hollowstar$ \\ 
Dependency Identification & $\solidstar\solidstar\solidstar$ & $\solidstar\hollowstar\hollowstar$ & $\hollowstar\hollowstar\hollowstar$ & $\hollowstar\hollowstar\hollowstar$ & $\hollowstar\hollowstar\hollowstar$ \\ 
Cross-Domain Knowledge & $\hollowstar\hollowstar\hollowstar$ & $\solidstar\solidstar\solidstar$ & $\hollowstar\hollowstar\hollowstar$ & $\hollowstar\hollowstar\hollowstar$ & $\solidstar\solidstar\solidstar$ \\ 
Domain-Specific Knowledge  & $\hollowstar\hollowstar\hollowstar$ & $\solidstar\solidstar\solidstar$ & $\hollowstar\hollowstar\hollowstar$ & $\hollowstar\hollowstar\hollowstar$ & $\solidstar\hollowstar\hollowstar$ \\ 
\hline
\end{tabular}
}
\caption{Constraining five fundamental task complexities (top) to construct test tasks across ten capability dimensions (left).}
\label{tab:cap}
\end{table}

\subsection{Experiment Setup}
\subsubsection{Settings}
For the evaluation phase, we follow the practices of works such as Aguvis and UGround, ensuring consistency in preprocessing and maintaining comparability across different models. Specifically, we standardize the image resolution by scaling all input images to \(1024 \times 1024\) pixels. This resolution is chosen to balance computational efficiency and visual detail, ensuring that models can effectively process graphical user interfaces (GUIs) without excessive memory overhead or loss of important information. The decision to use a fixed resolution is motivated by several key factors. First, many modern multimodal models and virtual agents, including those designed for GUI interaction, are trained on datasets with varying resolutions. Standardizing the input resolution reduces inconsistencies that may arise due to different input scales, helping models generalize better across diverse GUI layouts. Additionally, this resolution aligns with commonly used image sizes in recent benchmarks, facilitating direct comparison with prior works. Moreover, scaling images to \(1024 \times 1024\) ensures that finer details within the GUI, such as text labels, buttons, and icons, remain distinguishable without introducing excessive noise or aliasing effects. Given that GUI elements often contain intricate visual cues, a resolution that is too low may result in information loss, whereas an excessively high resolution can increase computational costs without significant performance benefits. To implement this resizing, we employ bicubic interpolation, which provides a good balance between sharpness and smoothness, preserving critical GUI features while minimizing artifacts. We also ensure that aspect ratios are maintained whenever possible by applying padding techniques if necessary, preventing unintended distortions that could affect model predictions. By adopting this resolution standard, we align our evaluation methodology with established works like Aguvis and UGround, fostering reproducibility and enabling fair comparisons in GUI-based task assessments.

\subsubsection{Baselines}
\label{baseline}
We conducted a comprehensive evaluation of the following four types of models: 

\textbf{1)~Closed-source MLLMs}: These include GPT-4o~\cite{gpt}, Qwen-VL-Max~\cite{qwenvl}, Claude-3.5-Sonnet, and Gemini-2.0-Flash, which are proprietary models developed by leading organizations. These models are selected due to their state-of-the-art performance on a wide range of multimodal tasks, including vision-language understanding, reasoning, and instruction following. 
Closed-source MLLMs typically benefit from large-scale pretraining on extensive proprietary datasets, incorporating a mixture of text and images/videos, along with advanced optimization techniques. They often leverage reinforcement learning from human feedback (RLHF) and continual updates to enhance their reasoning capabilities. These models are accessible via APIs, which enforce constraints on inference settings such as token limits, response latency, and proprietary decoding strategies. 
For our evaluation, we use their publicly available APIs and follow their recommended inference settings. If the model does not provide a specific prompt for agent tasks, we apply our unified prompt to ensure a fair comparison. Additionally, since these models lack direct access to GUI-specific training data, we evaluate their adaptability by providing structured information about the interface components.

\textbf{2)~Open-source MLLMs}: We also evaluate open-source models such as Qwen2-VL-7B-Instruct~\cite{qwen2vl}, InternVL2-8B~\cite{internvl}, and InternVL2.5-8B~\cite{internvl}, which are widely recognized in the research community for their flexibility and strong performance. These models have been fine-tuned on large-scale multimodal datasets and provide strong generalization across diverse vision-language tasks.
Open-source MLLMs offer several advantages, including transparency in training methodologies, customizability for domain-specific fine-tuning, and community-driven improvements. Unlike closed-source models, researchers can inspect their architectures, modify their training pipelines, and deploy them on local hardware without API restrictions. 
We use their fine-tuned weights and apply our unified prompt for tasks where no default prompt is provided. Although some of these MLLMs demonstrate capabilities in object recognition and grounding, directly instructing them to locate elements on graphical user interfaces (GUIs) presents unique challenges. The subtle and abstract nature of GUI elements, which differ significantly from natural objects in common vision datasets, makes accurate interpretation difficult. To mitigate this issue, we provide A11Y (accessibility) metadata, such as element descriptions and structural information, to aid inference.

\textbf{3)~Virtual Agents}: In addition to MLLMs, we also evaluate various virtual agents, such as Aguvis~\cite{aguvis}, OS-Atlas~\cite{os-atlas}, and ShowUI~\cite{showui}. These agents are designed for executing tasks and represent the current state-of-the-art in the virtual agent domain.
Unlike MLLMs, which primarily operate as general-purpose multimodal models, virtual agents are often specialized for task execution. They integrate structured knowledge representations, planning mechanisms, and fine-tuned models tailored to interactive environments. Many virtual agents leverage reinforcement learning or hierarchical decision-making frameworks to enhance their task completion efficiency. Some also incorporate retrieval-based techniques to access domain-specific knowledge dynamically.
Similar to the MLLMs, we use the default prompts provided by each agent when available. For agents that do not specify prompts for certain tasks, we apply our unified prompt to maintain consistency across experiments. Given that virtual agents typically rely on structured inputs rather than free-form multimodal understanding, we evaluate their ability to process GUI environments by simulating real-world task execution scenarios.

\textbf{4)~Supervised Fine-Tuning Agents}: We selected two backbones, OS-Atlas-Base-4B~\cite{os-atlas} and UGround-V1-7B~\cite{uground}, with different architectures and fine-tuned them using the synthetic data from OmniBench.
Supervised fine-tuning agents differ from pre-trained MLLMs and general-purpose virtual agents in that they are explicitly optimized on curated datasets. The fine-tuning process involves exposing the models to task-specific examples, allowing them to internalize structured dependencies and improve generalization within the OmniBench framework. This approach enables agents to learn robust action sequences, refine their perception of GUI elements, and enhance their decision-making accuracy.
By leveraging synthetic data, we ensure that these agents develop a structured understanding of GUI tasks while minimizing biases inherent in real-world datasets. We analyze their performance across varying task complexities, measuring improvements in execution success rates, response coherence, and adaptability to unseen scenarios.

\subsection{Evaluation Details}
In this section, we outline the evaluation strategies adopted for different categories of models, considering their architectural characteristics and observed behaviors.

For open-source MLLMs, we observe that models such as Qwen-VL exhibit limited ability to process high-resolution content across multiple images, often resulting in severe hallucinations. To mitigate this, we simplify the visual input by providing only a single high-resolution screenshot of the current state. This strategy reduces visual ambiguity and helps focus the model’s attention on the current interaction context.

In contrast, closed-source MLLMs, which typically benefit from larger model capacities and more extensive training data, demonstrate stronger capabilities in distinguishing differences across images. For these models, we provide two concatenated images: (1) the screenshot of the previous click location with red box annotations, and (2) the current high-resolution screenshot with similar markup. These two images are horizontally stitched into a single input to highlight temporal context and action history.

For OS-Atlas, we observe a consistent failure in zero-shot double-click interactions, even when the prompt explicitly emphasizes the importance of double-clicking. This limitation is particularly critical in the Windows environment, where double-clicking is often required for opening files or applications. To address this, we enforce all click actions as double-clicks for OS-Atlas during evaluation to ensure basic operability in such environments.

\section{Prompts}
In this section, we present the prompts used for constructing subtask trajectories and subtask evaluations.

\subsection{Prompt for Subtask Trajectory Synthesis}

\begin{tcolorbox}[
  colback=gray!5,
  colframe=blue!60!black,
  title=Example Prompt,
  listing only,
  breakable
]
\begin{minted}[
  fontsize=\footnotesize,
  breaklines=true,
  breakanywhere=true,
  breaksymbolleft={}, 
  breaksymbolright={}
]{yaml}
system: |-
  You are now operating in Executable Language Grounding mode. Your task is to help users accomplish their goals by suggesting executable actions based on the provided task instructions and your observations of the current situation. 

  ## Environment Interaction Rules

  ### Screenshots
  - You are provided two versions of screenshots of the current application in a single image, one with annotation (right) and one without annotation (left)
  - The annotation is to help you identify the control elements on the application
  - The annotation is a small rectangle with a number in the center of the rectangle in the top left corner of the control item. The number is the label of the control item
  - Different types of control items have different colors of annotation
  ### Control Items
  - The control item is the element on the page that you can interact with, such as button, input box, etc.
  - You are given the information of all available control items in the current application window in a list format:
    {{
      "label": <The annotated label of the control item>,
      "control_text": <The text of the control item>,
      "control_type": <The type of the control item>,
      "parent_control_text": <The text of the parent control item. When you are not sure which control to select, you can make a decision based on their parent controls>,
      "parent_control_type": <The type of the parent control item. When you are not sure which control to select, you can make a decision based on their parent controls>
    }}
  ### Control Operations
  - You are able to use pywinauto to interact with the control item
  {apis}

  ### Execution Status
  - You are required to determine the status of the execution after performing the current action. Choose from the following options and fill in the "Status" field in the response:
    - "CONTINUE": means the task is not yet completed and further actions are required. This is typically chosen when the execution is still ongoing or needs additional steps.
    - "FINISH": means the task has been fully completed, and all necessary actions have been carried out successfully. Only choose this when all steps have been executed as planned and the task is considered finished.
  
  ### Other Guidelines
  - You are required to respond in a JSON format, consisting of 8 distinct parts with the following keys and corresponding content:
    {{
      "Status": <Specify the status of the exploration. If "Status" is "FINISH", the "ControlLabel", "ControlText", "Function", and "Args" should be empty>,
      "Observation": <summarize the screenshot from the previous step, if it exists.  You can also compare the current screenshot with the one taken at the previous step>,
      "Thought": <Outline your thinking and logic of the current one-step action required to seek inspiration for task design>,
      "ControlLabel": <Specify the precise annotated label of the control item to be selected, adhering strictly to the provided options in the field of "label" in the control information. If you believe none of the control items are suitable for the task or the task is complete, kindly output an empty string ''>,
      "ControlText": <Specify the precise control_text of the control item to be selected, adhering strictly to the provided options in the field of "control_text" in the control information. If you believe none of the control items are suitable for the task or the task is complete, kindly output an empty string ''. The control text must match exactly with the selected control label>,
      "Function": <Specify the precise API function name without arguments to be called on the control item to complete the user request, e.g., click_input. Leave it an empty string "" if you believe none of the API functions are suitable for the task or the task is complete>,
      "Args": <Specify the precise arguments in a JSON object format of the selected API function to be called on the control item to complete the user request, e.g., {{"button": "left", "double": false}}. Leave it an empty dictionary {{}} if the API does not require arguments, or you believe none of the API functions are suitable for the task, or the task is complete>,
    }}

  Make sure your answer is strictly in JSON format only, without other redundant text such as json header. Your output must be able to be parsed by json.loads(). Otherwise, it will crash the system and destroy the user's computer.

user: |-
  <Step History:> {action_history}
  <Available Control Item:> {control_item}
  <Task instruction:> {task_instruction}

\end{minted}
\end{tcolorbox}

\subsection{Prompt for Subtask Evaluation Synthesis}

\begin{tcolorbox}[
  colback=gray!5,
  colframe=blue!60!black,
  title=Example Prompt,
  listing only,
  breakable
]
\begin{minted}[
  fontsize=\footnotesize,
  breaklines=true,
  breakanywhere=true,
  breaksymbolleft={}, 
  breaksymbolright={}
]{yaml}
system: |-
  You are a coding assistant tasked with generating Python code to evaluate if a digital agent has successfully completed a specific task. You will receive a task description along with a set of APIs that you can use to check different actions or conditions that indicate task completion. Your goal is to write an evaluation function that returns True if the agent has successfully completed the task and False otherwise.

  ### Available APIs:
  ```python
  def check_mouse_clicks(text: str) -> bool: 
    """Checks if the mouse has clicked on the specified text.
    Parameters
    ---------
    text: str
        The text associated with the click.
    Returns
    ---------
    bool
        True if the mouse has clicked on the specified text, False otherwise.
    Examples
    ---------
    >>> # Evaluate if the agent has successfully set the picture 'envelope.png' as background
    >>> def evaluate_agent_task_completion():
    >>>     if not check_mouse_clicks(text='envelope.png'):
    >>>         return False
    >>>     if not check_mouse_clicks(text='set as background'):
    >>>         return False
    >>>     return True
    """

  def check_keyboard_types(text: str) -> bool: 
    """Checks if the keyboard has typed the specified text.
    Parameters
    ---------
    text: str
        The text to be typed.
    Returns
    ---------
    bool
        True if the keyboard has typed the specified text, False otherwise.
    Examples
    ---------
    >>> # Evaluate if the agent has successfully typed 'Hello, World!'
    >>> def evaluate_agent_task_completion():
    >>>     if not check_keyboard_types(text='Hello, World!'):
    >>>         return False
    >>>     return True
    """

  def check_file_exists(file_path: str) -> bool: 
    """Checks if the specified file exists.
    Parameters
    ---------
    file_path: str
        The path to the file to be checked.
    Returns
    ---------
    bool
        True if the file exists, False otherwise.
    Examples
    ---------
    >>> # Evaluate if the agent has successfully renamed 'cat.jpg' to 'cute cat.jpg'
    >>> def evaluate_agent_task_completion():
    >>>     if check_file_exists(file_path='C:/Users/user/Desktop/images/cat.jpg'):
    >>>         return False
    >>>     if not check_file_exists(file_path='C:/Users/user/Desktop/images/cute cat.jpg'):
    >>>         return False
    >>>     return True
    """

  def check_text_exists_via_ocr(text: str) -> bool: 
    """Checks if the specified text is present in the last screenshot using OCR (Optical Character Recognition).
    Parameters
    ---------
    text: str
        The text to be checked.
    Returns
    ---------
    bool
        True if the text is present in the last screenshot, False otherwise.
    Examples
    ---------
    >>> # Evaluate if the agent has successfully set the clock to '9:00 AM'
    >>> def evaluate_agent_task_completion():
    >>>     if not check_text_exists_via_ocr(text='9:00 AM'):
    >>>         return False
    >>>     return True
    """

  def check_text_exists_via_control(text: str) -> bool: 
    """Checks if the specified text is present in the last screenshot through control information.
    Parameters
    ---------
    text: str
        The text to be checked.
    Returns
    ---------
    bool
        True if the text is present in the last screenshot, False otherwise.
    Examples
    ---------
    >>> # Evaluate if the agent has successfully input the code 'print("Hello World!")'
    >>> def evaluate_agent_task_completion():
    >>>     if not check_text_exists_via_control(text='print("Hello World!")'):
    >>>         return False
    >>>     return True
    """

  def check_text_exists(text: str) -> bool: 
    """Checks if the specified text is included in the last screenshot.
    Parameters
    ---------
    text: str
        The text to be checked.
    Returns
    ---------
    bool
        True if the text is present in the last screenshot, False otherwise.
    Examples
    ---------
    >>> # Evaluate if the agent has successfully created a new folder named 'Project Files'
    >>> def evaluate_agent_task_completion():
    >>>     if not check_text_exists(text='Project Files'):
    >>>         return False
    >>>     return True
    """
  ```

  ### Other Guidelines
  - You will be given a `Subtask Instruction Template` and `Parameters`. Use the APIs provided to implement an `Evaluation Function` in Python.
  - This agent will run on the `Windows 11` operating system, so please consider how to cleverly design the evaluation function based on this operating system.
  - The evaluation function should return a namedtuple `EvalResult` with two fields:
    - `success`: A boolean indicating if all conditions are met (True) or not (False)
    - `message`: A string explaining why the evaluation succeeded or failed
  - The evaluation function should check each required condition and return appropriate success/failure messages.
  - Please `directly output` the evaluation function, without any additional comments or explanations.
  - When you design a correct evaluation function, I will provide you with a `$1000` tip.

user: |-
  ### Subtask Instruction Template
  {instruction}

  ### Available Parameters
  {parameters}

  ### Controls in Environment
  {controls}
  
\end{minted}
\end{tcolorbox}

\end{document}